\documentclass[10pt,twocolumn,letterpaper]{article}

\usepackage{cvpr}              

\usepackage{amsmath}
\usepackage{graphicx}
\usepackage{multirow}
\usepackage{wrapfig} 
\usepackage{makecell}
\usepackage{caption}
\usepackage{blindtext}
\usepackage{comment}
\usepackage{lipsum}
\usepackage{colortbl}
\usepackage[utf8]{inputenc} 
\usepackage[T1]{fontenc}    
\usepackage{url}            
\usepackage{booktabs}       
\usepackage{amsfonts}       
\usepackage{nicefrac}       
\usepackage{microtype}      
\usepackage{xcolor}         
\usepackage{algorithm}
\usepackage{algpseudocode}
\usepackage[accsupp]{axessibility}  

\usepackage[dvipsnames]{xcolor}

\definecolor{cvprblue}{rgb}{0.21,0.49,0.74}
\usepackage[pagebackref,breaklinks,colorlinks,allcolors=cvprblue]{hyperref}

\title{ITA-MDT: Image-Timestep-Adaptive Masked Diffusion Transformer Framework for Image-Based Virtual Try-On}

\author{
    Ji Woo Hong$^{1}$ \quad 
    Tri Ton$^{1}$ \quad 
    Trung X. Pham$^{1}$ \quad 
    Gwanhyeong Koo$^{1}$ \quad 
    Sunjae Yoon$^{1}$ \quad 
    Chang D. Yoo$^{1}$ \\
    $^{1}$Korea Advanced Institute of Science and Technology (KAIST), South Korea\\
    {\tt\small \{jiwoohong93, tritth, trungpx, kookie, dbstjswo505, cd\_yoo\}@kaist.ac.kr}
}

\begin{document}

\twocolumn[{
\maketitle
\begin{center}
   \includegraphics[width=1.0\textwidth]{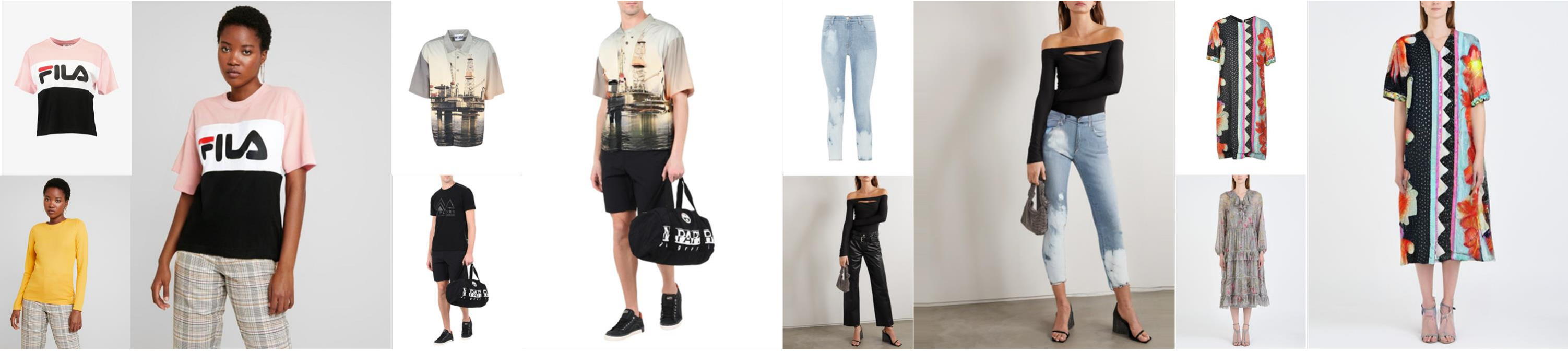}
   \captionof{figure}{Image-based Virtual Try-on results generated by our ITA-MDT. From left to right: VITON-HD\cite{choi2021viton}, DressCode\cite{morelli2022dress} upper, DressCode lower, and DressCode dresses, respectively.}
\label{fig:teaser}
\end{center}
}]

\begin{abstract}
This paper introduces ITA-MDT, the Image-Timestep-Adaptive Masked Diffusion Transformer Framework for Image-Based Virtual Try-On (IVTON), designed to overcome the limitations of previous approaches by leveraging the Masked Diffusion Transformer (MDT) for improved handling of both global garment context and fine-grained details. The IVTON task involves seamlessly superimposing a garment from one image onto a person in another, creating a realistic depiction of the person wearing the specified garment. Unlike conventional diffusion-based virtual try-on models that depend on large pre-trained U-Net architectures, ITA-MDT leverages a lightweight, scalable transformer-based denoising diffusion model with a mask latent modeling scheme, achieving competitive results while reducing computational overhead. A key component of ITA-MDT is the Image-Timestep Adaptive Feature Aggregator (ITAFA), a dynamic feature aggregator that combines all of the features from the image encoder into a unified feature of the same size, guided by diffusion timestep and garment image complexity. This enables adaptive weighting of features, allowing the model to emphasize either global information or fine-grained details based on the requirements of the denoising stage. Additionally, the Salient Region Extractor (SRE) module is presented to identify complex region of the garment to provide high-resolution local information to the denoising model as an additional condition alongside the global information of the full garment image. This targeted conditioning strategy enhances detail preservation of fine details in highly salient garment regions, optimizing computational resources by avoiding unnecessarily processing entire garment image. Comparative evaluations confirms that ITA-MDT improves efficiency while maintaining strong performance, reaching state-of-the-art results in several metrics. Our project page is available at \url{https://jiwoohong93.github.io/ita-mdt/}.

\end{abstract}

\section{Introduction}
\label{sec:intro}

The task of Image-based Virtual Try-on (IVTON) is to generate an image where a person appears dressed in specific clothes, using provided images of both the garment and the individual. 
In real-world applications, IVTON holds great potential for revolutionizing online shopping, fashion design \cite{de2023disentangling, hsiao2018creating, sarkar2023outfittransformer, hadi2015buy}, and even entertainment industries for creating digital avatars or virtual costumes \cite{ma2021pixel, prajwal2020lip, xu2024vasa}.
For this task, deep learning methods \cite{han2018viton, wang2018toward, issenhuth2020not, yang2020towards, choi2021viton, ge2021parser, lee2022high, xie2023gp} employ paired datasets \cite{choi2021viton, morelli2022dress} containing both standalone garment images and the images of the individual wearing these for training their models, along with additional reference images such as mask images of the garment region and DensePose \cite{guler2018densepose} image. 

Recently, diffusion models \cite{ho2020denoising, song2020denoising} have shown great promise in image generation and synthesis in many different tasks \cite{pham2024cross, yoon2024tpc, yoon2024dni, koo2024flexiedit, yoon2024frag}. Especially, large-scale pre-trained diffusion models \cite{ramesh2021zero, rombach2022high, saharia2022photorealistic}, such as Stable Diffusion \cite{rombach2022high}, are widely used to preserving high fidelity in generated images; however, their large size and slow inference speeds present practical limitations in commercial application.

To address this, transformer-based diffusion models have emerged as an efficient alternative. The Diffusion Transformer (DiT) \cite{peebles2023scalable} based on the Visual Transformer (ViT) \cite{dosovitskiy2020image} architecture with Latent Diffusion Models \cite{rombach2022high} can achieve excellent scalability and efficiency. Extending this concept, the Masked Diffusion Transformer (MDT) \cite{gao2023masked} applied mask prediction scheme to enhance spatial representation learning, achieving state-of-the-art results in class-conditional image generation. While promising, directly applying MDT to IVTON task presents challenges, as the task requires high fidelity and consistency across multiple reference images, demanding more than a straightforward approach.

This paper presents the ITA-MDT (Image-Timestep-Adaptive Masked Diffusion Transformer) Framework, which incorporates our modified version of MDT adapted specifically for IVTON, named MDT-IVTON, as well as two novel components: the Image-Timestep Adaptive Feature Aggregator (ITAFA) and the Salient Region Extractor (SRE). This framework is designed to preserve garment fidelity while maintaining computational efficiency.
Our MDT-IVTON model leverages MDT’s mask prediction scheme to efficiently learn spatial relationships between garment and individual. Additionally, we incorporate an auxiliary loss to enhance the model’s focus on the inpainting objective of the task. 
While this architecture effectively captures global dependencies, it struggles to retain fine-grained details, which U-Net-based models handle relatively naturally through their multi-scale encoder-decoder structure that captures and reconstructs information of the image across multiple scales through progressive downsampling and upsampling.
Our ITAFA and SRE bridge this gap.

\begin{figure}[t!]
\centering
   \includegraphics[width=\columnwidth]{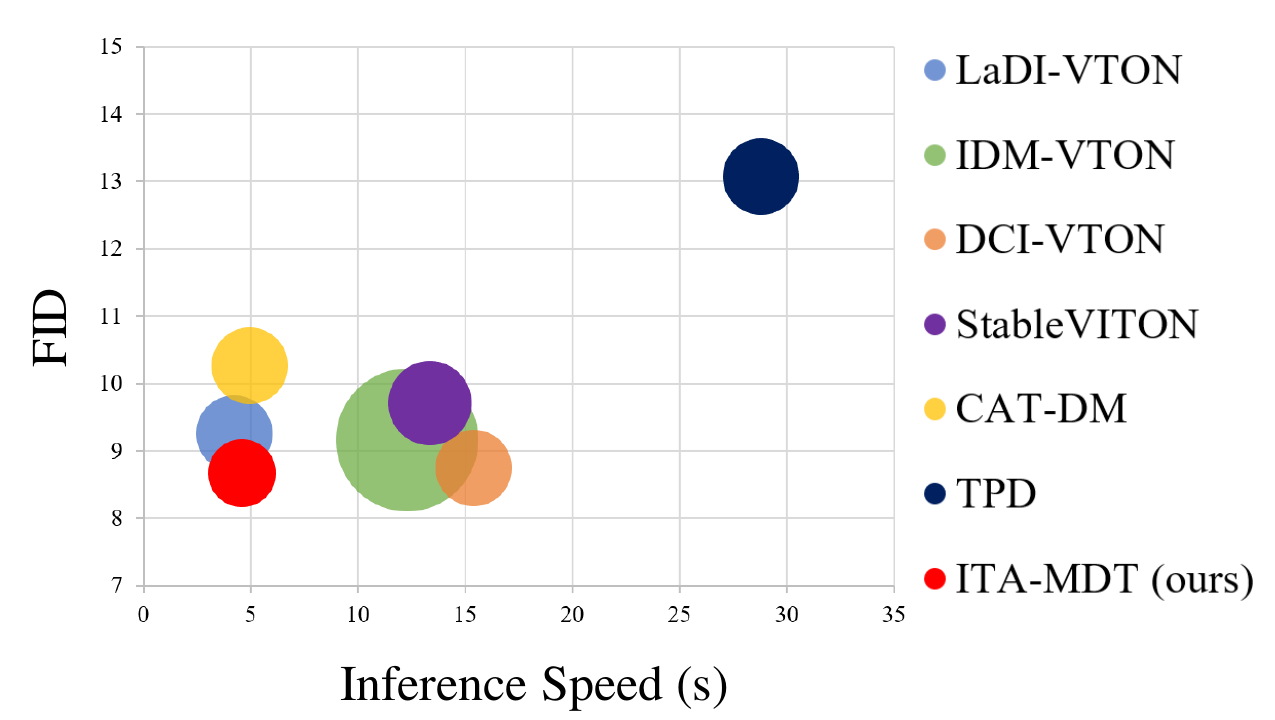}
   \caption{Comparison of our ITA-MDT with other diffusion-based methods on the VTON task using the VITON-HD dataset. For all metrics in the analysis, including FID, model parameters, and inference speed, lower values indicate better performance.}
\label{fig:efficiency}

\end{figure}

The Image-Timestep Adaptive Feature Aggregator (ITAFA) is a learnable feature aggregator designed to condition the diffusion model effectively and efficiently. It dynamically combines multiple image features from ViT image encoder, guided by the current diffusion timestep and garment image complexity score. 
While simply providing multiple image features from different layers can condition the model with extensive information, this naive approach significantly increases the computational load.
By assigning adaptive weights to each feature, ITAFA aggregates them into a single representation. These weights are modulated based on the timestep information, which indicates how far the generation process has progressed. The weighting is further refined using the visual complexity of the garment, which reflects how much fine-grained information the condition image contains. 
This design enables the model to gradually shift its focus from global semantic features in the early stages to fine local details in the later stages of the diffusion process.

The Salient Region Extractor (SRE) provides high-resolution local information of key garment area, such as logos or complex patterns, identified through entropy-based detection. The region extracted offers additional crucial condition guidance to the diffusion model alongside the global context from the full garment image.
While simply using a high-resolution, big size, image enhances detail representation, it results in substantial computation. 
Notably, most regions of garments are either solid-colored or display large, simple patterns that can be captured adequately through lower resolution.
Therefore, rather than uniformly processing the entire image at high resolution, SRE selectively extracts and provides additional high-resolution information only for regions where fine-grained local information is crucial. In summary, our contributions include:
\begin{enumerate}
\item We present ITA-MDT, a framework for the image-based virtual try-on (IVTON) task that builds upon MDT-IVTON, our adaptation of the Masked Diffusion Transformer (MDT) for IVTON, and incorporates our proposed methods to effectively address the challenges of transformer-based denoising diffusion models in preserving fine-grained details.

\item We propose ITAFA, a learnable feature aggregator designed to dynamically combine features from the image encoder into a single feature based on the current denoising timestep and garment image complexity score.  

\item We propose SRE to provide high-resolution local information of key garment area, identified through entropy-based detection, as additional condition guidance to the diffusion model alongside the global context from the full garment image.

\item ITA-MDT’s combination of MDT-IVTON, ITAFA, and SRE offers a highly efficient solution for IVTON, achieving a balance between model efficiency and output fidelity. 
Figure~\ref{fig:efficiency} presents the comparison of our ITA-MDT with previous diffusion-based methods. 
It achieves state-of-the-art results in several metrics while maintaining competitive performance across the others, demonstrating its overall effectiveness and suitability for practical virtual try-on applications.

\end{enumerate}

\section{Related Work}

\subsection{Image-based Virtual Try-on with Diffusion}
The integration of diffusion models has significantly advanced the field of Image-based Virtual Try-on, as evidenced in numerous recent studies. Notably, TryOnDiffusion \cite{zhu2023tryondiffusion} introduced an architecture utilizing dual U-Nets, highlighting the potential of diffusion-based try-on but requiring large-scale datasets composed of image pairs of the same person in varied poses. This large-scale dataset requirement has led researchers to pivot towards leveraging large pre-trained diffusion models \cite{ramesh2021zero, rombach2022high, saharia2022photorealistic}.
To condition these models effectively, methods such as representing garments as pseudo-words in LaDI-VTON \cite{morelli2023ladi}, applying warping networks in DCI-VTON \cite{gou2023taming}, modifying attention mechanisms in StableVITON \cite{kim2024stableviton} and IDM-VTON \cite{choi2024improving}, utilizing ControlNet for garment control and GAN-based sampling in CAT-DM \cite{zeng2024cat}, and concatenating masked person and garment images for direct texture transfer in TPD \cite{yang2024texture} have been introduced.
However, the use of large pre-trained U-Nets comes with notable limitations. 
The large number of parameters in these models significantly increases memory usage and slows down inference speed, making them resource-intensive and less practical for efficient deployment. A variety of solutions could be explored to address these issues, among which transformer-based diffusion models present a promising approach.

\subsection{Transformers for Diffusion Process}
Traditionally, The U-Net architecture \cite{ronneberger2015u} has been foundational for diffusion models and continues to be a prevalent choice across various diffusion-based generation tasks \cite{ho2020denoising, song2019generative}. However, the incorporation of transformers into diffusion processes also marked a significant evolution in generative modeling. 
The introduction of the Diffusion Transformer (DiT) \cite{peebles2023scalable} merged the architecture of the Vision Transformer (ViT) \cite{dosovitskiy2020image} with latent diffusion models. 
DiT has demonstrated outstanding scalability and compute efficiency, delivering superior generative performance compared to traditional U-Net architectures. 
Building on this, the Masked Diffusion Transformer (MDT) \cite{gao2023masked} has further advanced the diffusion transformer model by employing an asymmetric masking schedule to optimize the contextual representation learning, setting a new benchmark in class image generation on ImageNet. 
In an attempt to leverage MDT for image-to-image generation, \cite{pham2024cross} introduced an aggregation network that consolidates image conditions into a single vector for guiding the diffusion process in pose-guided human image generation tasks. While effective within its domain, directly applying this approach to the IVTON task is unsuitable. IVTON involves a significantly larger set of condition images, and compressing these into a single vector leads to substantial information loss, ultimately degrading output quality. Given the need to preserve fine details from condition images, enhancing rather than compressing condition information is preferred.

\subsection{Enhancing Condition Information in Virtual Try-On task}
To enhance the intricate details essential to IVTON, the recent method, PICTURE \cite{ning2024picture}, introduces a two-stage disentanglement pipeline that separates style and texture features by employing a hierarchical CLIP feature extraction module with position encoding, enabling the representation of complex, non-stationary textures. While effective in handling intricate style and texture combinations, this method requires an in-depth analysis of features from the hidden layers of the image encoder. Specifically, this involves clustering hidden layer features and manually selecting representative features from each cluster, a process that lacks flexibility for it requires re-evaluation when a different encoder is to be used. Additionally, the selected features from each cluster are concatenated for cross-attention conditioning, substantially increasing computational demand. Together, these requirements render the approach resource-intensive and less adaptable to diverse architectures.

\section{Method}
\begin{figure*}[ht!]
\centering
   \includegraphics[width=1.0\textwidth]{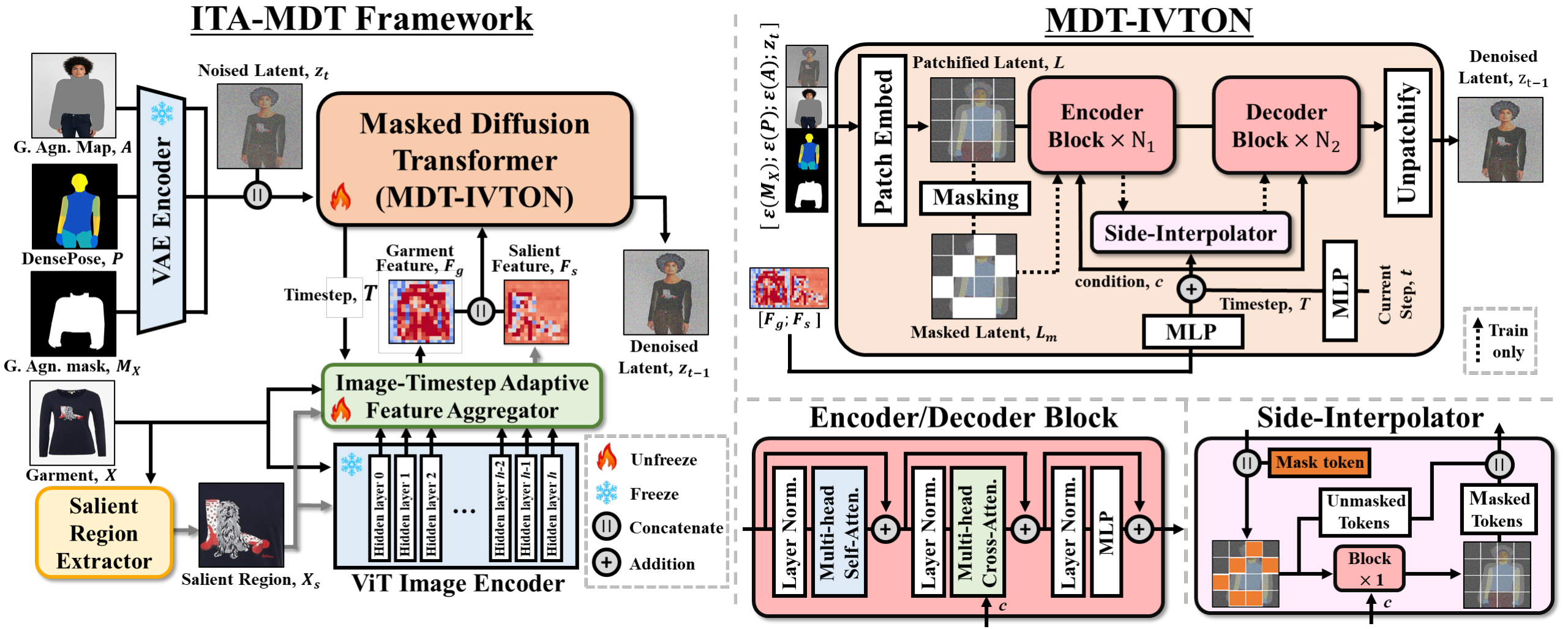}
   \caption{
   Overview of the ITA-MDT Framework for Image-Based Virtual Try-On (IVTON). The framework takes multiple reference images encoded into latent space as query, which includes Garment Agnostic Map $A$, DensePose $P$, and Garment Agnostic Mask $M_X$. These reference latent images are concatenated to be patch embeded for masked diffuison process within our MDT-IVTON, which follows the architecture of MDTv2 \cite{gao2023masked} with integrated cross-attention blocks. The image feature of Garment $X$ is extracted with the ViT image encoder, DINOv2 \cite{oquab2023dinov2}, and are adaptively aggregated with our proposed Image-Timestep Adaptive Feature Aggregator (ITAFA) to produce Garment Feature $F_g$. With our Salient Region Extractor (SRE), the Salient Region $X_s$ is extracted from the Garment $X$ and processed through ITAFA separately to produce Salient Feature $F_s$. The Garment Feature $F_g$ and Salient Feature $F_s$ are concatenated to serve as conditions of MDT-IVTON. Positional embeddings are omitted in this figure for simplicity.
   }

\label{fig:overview}
\end{figure*}

The ITA-MDT framework performs Image-Based Virtual Try-on through a transformer-based denoising diffusion network, leveraging the mask latent modeling scheme of MDTv2 \cite{gao2023masked}. An overview of the architecture is shown in Figure \ref{fig:overview}, which comprises three primary components: (1) the Masked Diffusion Transformer for Image-Based Virtual Try-On (MDT-IVTON), (2) the Image-Timestep Adaptive Feature Aggregator (ITAFA), and (3) the Salient Region Extractor (SRE) module. Detailed descriptions of each component are provided below.

\subsection{Masked Diffusion Transformer for Image-Based Virtual Try-On (MDT-IVTON)}
\label{sec:mdt-ivton}

Our ITA-MDT framework integrates a Masked Diffusion Transformer (MDT) tailored for Image-Based Virtual Try-On (IVTON), leveraging the transformer-based denoising diffusion model architecture of DiT \cite{peebles2023scalable}. This model operates in a latent space encoded via a pre-trained VAE following the Latent Diffusion Model (LDM) approach \cite{rombach2022high}. Specifically, the VAE encoder maps garment agnostic inputs into this latent space, including the Garment Agnostic Map \( A \), DensePose \( P \), and Garment Agnostic Mask \( M_X \), concatenated with the noised latent representation \( z_t \).

The MDT-IVTON model processes the input by embedding it into a patchified latent representation \( L \) and sequentially passing it through $N_1$ Encoder Blocks followed by $N_2$ Decoder Blocks. Each block utilizes Layer Normalization, Multi-Head Self-Attention, Multi-Head Cross Attention, Multi-Layer Perceptron (MLP), and skip connections to refine the denoised latent representation \( z_{t-1} \), as illustrated in the right side of Figure \ref{fig:overview}. Conditioning information derived from the concatenated garment features \( F_g \) and \( F_s \), extracted from garment image \( X \) and its salient region \( X_s \) using the ViT encoder and ITAFA module, is infused through cross-attention mechanisms to guide the generation process. Detailed descriptions of ITAFA and SRE are available in \cref{sec:itafa} and \cref{sec:sre}.

Three objective functions guide the MDT-IVTON training to optimize its generative performance:

\subsubsection{Denoising Objective}
The primary objective function minimizes the mean squared error (MSE) between the predicted noise and the actual noise in the denoised latent \( z_t \) at each timestep \( t \), following the standard diffusion objective:
\begin{equation}
L_{\text{denoise}} = \mathcal{E}_{y, c, \epsilon \sim \mathcal{N}(0, I), t} \left[\lVert \epsilon - \epsilon_{\theta}(y_t, c, t)\rVert^2 \right],
\end{equation}
where \( y_t \) denotes the latent representation at timestep \( t \), \( c \) the conditioning, and \( \epsilon_{\theta} \) the model's predicted noise.

\subsubsection{Mask Reconstruction Objective}
In MDT-IVTON, the Mask Reconstruction Objective enforces spatial coherence and semantic consistency by reconstructing masked tokens within the masked latent representation \( E_m \). This objective ensures alignment between the reference and generated images, enhancing the model’s ability to synthesize realistic outputs in masked regions.

To achieve this, we incorporate the Side-Interpolator module, which plays a pivotal role in interpolating information across unmasked tokens to predict values for masked tokens in \( E_m \). The Side-Interpolator effectively fills in missing information by conditioning on neighboring unmasked regions, encouraging consistent token-level synthesis within the masked latent.

The Mask Reconstruction Objective follows an MSE formulation similar to the denoising objective:
\begin{equation}
L_{\text{mask}} = \mathcal{E}_{y, c, \epsilon \sim \mathcal{N}(0, I), t} \left[\lVert \epsilon - \epsilon_{\theta}(\textbf{Side-Int}(E_m), c, t)\rVert^2 \right],
\end{equation}
where \textbf{Side-Int} processes \( E_m \), leveraging spatial context to predict masked regions. The conditioning \( c \) includes garment details, aiding in preserving fidelity to the garment appearance in masked areas. This objective promotes detailed reconstruction by interpolating across the context in \( E_m \), enhancing semantic alignment within generated images.

\subsubsection{Inpainting Objective}
The inpainting objective in MDT-IVTON is designed to enhance garment fidelity by concentrating on regions defined by the garment agnostic mask \( M_X \), encoded as \( \mathcal{E}(M_X) \) in latent space. During training, this mask serves as a selective weighting factor in the mean squared error (MSE) between the target and predicted outputs, guiding the model’s focus to reconstruct details more accurately within garment-specific regions.

To ensure numerical stability and maintain well-behaved gradients, the mask is normalized to the range \([0, 1]\), preventing extreme values that could otherwise destabilize training. This targeted inpainting strategy enhances garment detail fidelity by directing the model’s attention to areas where high-quality reconstruction is most critical.

\textbf{Latent Mask Normalization}: To prevent unstable gradients, the single channel of the latent mask \( \mathcal{E}(M_X) \) that most closely resembles the binary mask of the inpainting task is chosen to be used as the mask across all channels in the loss calculation. This selected channel is normalized to values between 0 and 1 to ensure stability:
\begin{equation}
\mathcal{E}(M_X) = \frac{\mathcal{E}(M_X) - \min(\mathcal{E}(M_X))}{\max(\mathcal{E}(M_X)) - \min(\mathcal{E}(M_X)) + 1 \times 10^{-8}},
\end{equation}
where \( 1 \times 10^{-8} \) ensures numerical stability. This approach avoids unintended gradients and aligns the loss with garment-focused regions. A visualization of the latent mask selected is in 
the Supplementary Materials.

\textbf{Masked Loss Computation}: With the rescaled single channel of \( \mathcal{E}(M_X) \), the masked inpainting loss is calculated as a weighted mean squared error (MSE) applied uniformly across all channels in the latent space:
\begin{equation}
L_{\text{inpaint}} = \mathcal{E}_{y, c, t} \left[\lVert \mathcal{E}(M_X) \cdot (y_t - y_{t-1}) \rVert^2 \right],
\end{equation}
where \( y_t \) and \( y_{t-1} \) denote the latent representations at timesteps \( t \) and \( t-1 \), respectively. This loss term emphasizes reconstruction quality in the masked regions, promoting accurate garment-specific detail representation. Further details of MDT-IVTON are provided in 
Supplementary Materials.


\begin{figure*}[t!]
\centering
   \includegraphics[width=1.0\textwidth]{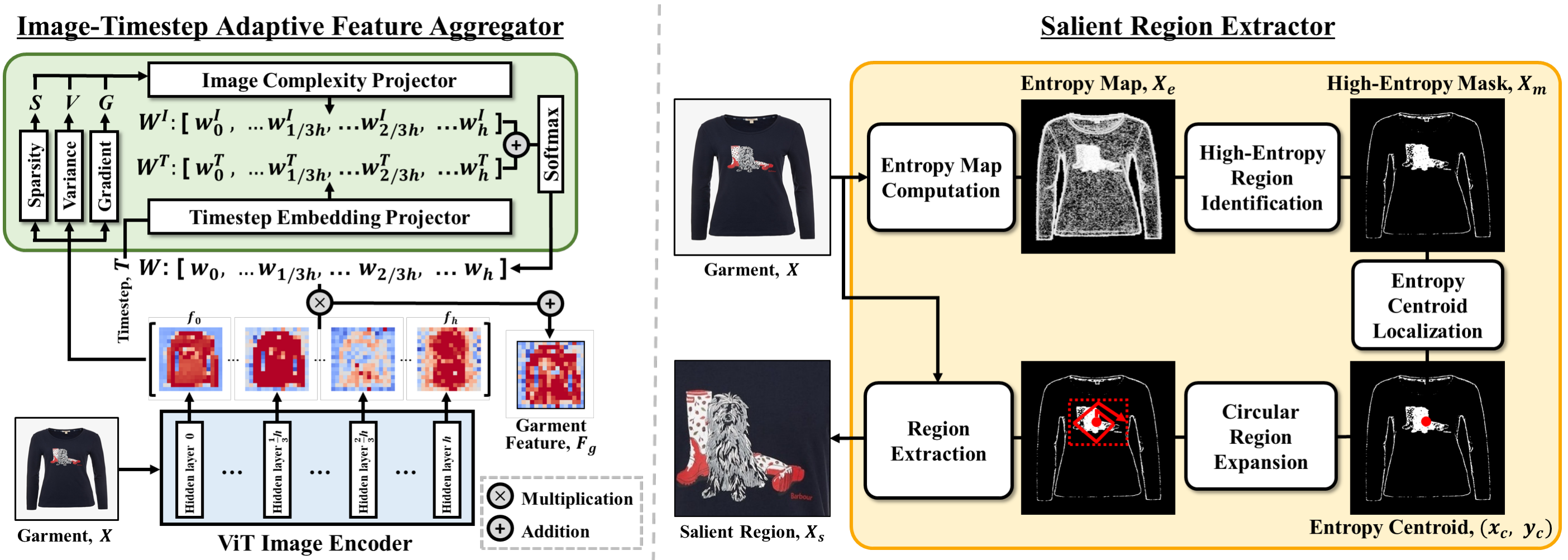}
    \caption{Overview of the Image-Timestep Adaptive Feature Aggregator (ITAFA) and the Salient Region Extractor (SRE). 
    \textbf{Left:} The ITAFA dynamically aggregates Vision Transformer (ViT) feature embeddings $f$ based on a combination of the Timestep Embedding Projector, which projects the diffusion timestep embedding $T$ to match the feature embedding dimensions, and the Image Complexity Projector, which transforms the image complexity vector $[S, V, G]$ (sparsity, variance, gradient magnitude) into a comparable dimension. The weight vectors are combined and normalized via softmax to form $W$, which is used to adaptively aggregate the feature embeddings $\{ f_i \}_{i=0}^h$ across hidden layers to produce the final output tensor $F$. 
    Garment $X$ and Salient Region $X_s$ are processed separately through ITAFA to generate the Garment Feature $F_g$ and Salient Feature $F_s$. 
    \textbf{Right:} The SRE processes the input image $X$ by computing an entropy map $X_e$, creating a binary High-Entropy Mask $X_m$, and circular region expansion from the entropy centroid to extract the final high-entropy region $X_s$, ensuring preservation of detail within a consistent aspect ratio.} 

\label{fig:itafa_sre}
\end{figure*}

\subsection{Image-Timestep Adaptive Feature Aggregator (ITAFA)}
\label{sec:itafa} 
The ITAFA is a dynamic feature aggregation module designed to adaptively weight feature embeddings from a Vision Transformer (ViT) image encoder, utilizing learnable parameters and feature complexity analysis to dynamically modulate the contribution of each hidden layer output of the image encoder at different timesteps of the denoising diffusion process.

\subsubsection{Components and Parameters}
\paragraph{Timestep Embedding Projector.} The diffusion timestep embedding, denoted as $T$, represents the current timestep in the denoising process with dimension $\mathcal{H}$ of the MDT-IVTON model. To align $T$ with the image feature dimension, it is projected to the dimensionality of the hidden layer of the ViT, $\mathcal{H_E}$. This is achieved through a linear transformation:
\begin{equation}
W^T = \text{Linear}(\mathcal{H}, \mathcal{H_E}),
\end{equation}
The linear layer maps the timestep embedding into a representation suitable for interacting with the ViT features.

\paragraph{Image Complexity Projector.} ITAFA incorporates three complexity components: feature sparsity, feature variance, and gradient magnitude. These collectively capture the complexity of garment texture and structure.
\begin{enumerate}
    \item \textbf{Feature Sparsity} measures the proportion of near-zero activations in the feature embedding tensor $f$, providing insight into the sparsity of learned representations. It is computed as:
    \begin{equation}
    \text{S} = \frac{1}{n} \sum_{i=1}^{n} \mathbb{I}(|f_i| < \delta),
    \end{equation}
    where $\delta$ is a small threshold (i.e., 0.01), $n$ denotes the total number of elements in $f$, and $\mathbb{I}(\cdot)$ is the indicator function.

    \item \textbf{Feature Variance} captures the variability of activations, reflecting structural complexity within the image:
    \begin{equation}
    V = \text{Var}(f),
    \end{equation}
    where $\text{Var}(f)$ denotes the variance across all elements in the feature embedding.

    \item \textbf{Gradient Magnitude} measures local changes in the feature embedding, calculated by taking the spatial gradients along patch and embedding dimensions. The magnitude of gradients are computed, and averaging them provides a single score $G$ that captures texture complexity.
    
    \begin{equation}
    G = \frac{1}{n} \sum_{i=1}^{n} \sqrt{\left(\frac{\partial f}{\partial x}\right)^2 + \left(\frac{\partial f}{\partial y}\right)^2}.
    \end{equation}

\end{enumerate}
These components are concatenated to form the complexity score vector $[S, V, G]$, which is then linearly projected to match the hidden layer dimension $\mathcal{H_E}$:
\begin{equation}
W^I = \text{Linear}(3, \mathcal{H_E}).
\end{equation}

\paragraph{Learnable Weighting Parameter.} The contribution of the timestep and complexity projections is balanced by a learnable parameter $\alpha \in [0, 1]$, initialized to 0.5 and optimized during training:
\begin{equation}
\label{eq_alpha}
W^{com} = \alpha \cdot W^T + (1 - \alpha) \cdot W^I.
\end{equation}
This weighted combination provides an adaptive control over the influence of timestep and image complexity on the image feature aggregation.

\subsubsection{Feature Aggregation}

The aggregated weights are normalized via softmax to ensure they sum to one across hidden layers, thereby controlling the contribution of each layers.
\begin{equation}
W = \text{softmax}(W^{com}).
\end{equation}
These weights are then applied to each hidden layer output from the ViT encoder, denoted by $f$, via an element-wise multiplication and summation across layers:
\begin{equation}
F = \sum_{l=1}^{\mathcal{H}} W[l] \cdot f[l],
\end{equation}
where $F$ represents the final aggregated feature tensor, capturing a balanced representation based on both denoising timestep and image complexities.

In-depth details of ITAFA are explained in 
Supplementary Materials.


\subsection{Salient Region Extractor (SRE)}
\label{sec:sre}
Our SRE is a simple yet effective algorithm designed to identify and extract salient, high-entropy, areas within an image. The method includes: Entropy Map Computation, High-Entropy Region Identification, Entropy Centroid Localization, Circular Region Expansion, and Region Extraction. 

By utilizing Shannon Entropy \cite{shannon1948mathematical} to compute an entropy map $X_e$, which measures information content across pixel neighborhoods. High-entropy regions are then identified by thresholding this map to create a binary High-Entropy Mask $X_m$ that highlights areas with substantial visual complexity. The centroid of this region is computed to serve as the initial center for region extraction. From this centroid, a circular region expansion method grows the bounding box outward in small increments, checking for low-entropy areas along each edge as stopping point to ensure meaningful content within the boundary. After the expansion, the region is expanded to follow the aspect ratio of the original image, then extracted as the final output $X_s$, capturing detailed areas while maintaining minimal distortion.

Details of SRE process and algorithm are elaborated in
Supplementary Materials.
\section{Experiments}

\begin{table*}[htbp]
    \centering
    \begin{tabular}{l|cc|cccccccc}
        \toprule
        & \multicolumn{1}{c}{Inf. Time} & \multicolumn{1}{c|}{Param.} & \multicolumn{3}{c}{VITON-HD} & \multicolumn{3}{c}{DressCode Upper} \\
        Method & (s)↓ & (M)↓ & LPIPS↓ & SSIM↑ & FID(pair/unp.)↓ & LPIPS↓ & SSIM↑ & FID(pair/unp.)↓ \\
        \midrule

        LaDI-VTON \cite{morelli2023ladi} &\textbf{4.225} & 865 & 0.098 & 0.881 & 6.587/9.265 & 0.062 & 0.927 & 10.447/14.048 \\
        DCI-VTON \cite{gou2023taming} & 15.408 & \underline{859} & 0.085 & \textbf{0.896} & \underline{5.559}/\underline{8.750} & 0.068 & 0.918 & 10.501/12.490 \\
        StableVITON \cite{kim2024stableviton} & 13.357 & 1,039 & 0.086 & 0.872 & 7.039/9.717 & 0.066 & 0.919 & 11.572/13.666 \\
        CAT-DM \cite{zeng2024cat} & 4.947 & \underline{859} & 0.104 & 0.855 & 7.697/10.263 & 0.058 & 0.922 & 10.257/14.039 \\
        TPD \cite{yang2024texture} & 28.784 & \underline{859} & 0.097 & 0.885 & 10.161/13.071 & 0.048 & \textbf{0.956} & 9.500/12.904 \\
        IDM-VTON \cite{choi2024improving} & 12.288 & 2,991 & \textbf{0.082} & 0.881 & 6.000/9.163 & \underline{0.036} & 0.944 & 7.471/\underline{11.915} \\
        
        \midrule
        MDT-IVTON (Base) & \underline{4.596} & \textbf{671} & 0.099 & \underline{0.887} & 7.773/9.838 & 0.061 & 0.921 & 10.300/13.998 \\
        Base+ITAFA & \underline{4.596} & \textbf{671} & 0.086 & 0.881 & 5.717/8.879 & \textbf{0.034} & \underline{0.951} & 5.414/13.111 \\
        Base+ITAFA+HR & 6.215 & \textbf{671} & \underline{0.083} & 0.884 & 5.542/8.800 & \textbf{0.034} & \underline{0.951} & \textbf{5.384}/12.528 \\
        Base+ITAFA+SRE & \underline{4.596} & \textbf{671} & \underline{0.083} & 0.885 & \textbf{5.462}/\textbf{8.676} &\textbf{0.034} & \underline{0.951} & \underline{5.412}/\textbf{10.069} \\ 
        \bottomrule
    \end{tabular}
    \caption{Quantitative results of diffusion-based virtual try-on methods on VITON-HD and DressCode upper-body test sets. Inf. Time refers to the inference time, diffusion model only, measured for generating a single image on our environment, and Param. refers to the denoising diffusion model's parameter size. We report both paired and unpaired FID. Following the previous works, feature of last hidden layer of ViT image encoder is used for our MDT-IVTON without ITAFA. HR refers to use of single High-resolution $(448 \times 448 \times 3)$ garment image to formulate condition vector $c$. The best performance for each metric is indicated in \textbf{bold}, while the second best is \underline{underlined}.
    }

    \label{tab:qual_table}
\end{table*}

\textbf{Implementation details}: 
\label{sec:imp_details} 
We use the MDTv2 XL model \cite{gao2023masked} with RGB reference images of shape \( 512 \times 512 \times 3 \), encoded into latent \( Z \)-space via the Stable Diffusion XL VAE \cite{podell2023sdxl}. These latent features are patch-embedded and processed within our model. For conditioning of garment features, the ViT encoder extracts features from RGB images of shape \( 224 \times 224 \times 3 \). Training was conducted on two 80GB A100 GPUs with a batch size of 6. Further details are provided in the Supplementary Materials.

\textbf{Datasets and Evaluation Metrics}: 
\label{sec:data_metric_details} 
We train and evaluate our model on two public datasets: VITON-HD \cite{choi2021viton} and DressCode \cite{morelli2022dress} upper, lower, and dresses. Data augmentation techniques from StableVITON \cite{kim2024stableviton} are applied to enhance model robustness.
To assess performance, we use standard metrics from previous VTON works: LPIPS \cite{zhang2018unreasonable}, SSIM \cite{wang2004image}, and FID \cite{heusel2017gans}. 
All evaluations follow the protocol established in LaDI-VTON \cite{morelli2023ladi} to ensure fair comparison with previous virtual try-on models.

Further information about the datasets and evaluation metrics is provided in
the Supplementary Materials.

\subsection{Qualitative and Quantitative Results}
The quantitative Results of Table. \ref{tab:qual_table} compare our proposed ITA-MDT framework to previous diffusion-based methods, as well as show effect of ITAFA, and SRE on MDT-IVTON. The values of the previous works are re-measured by implementation with their checkpoints released. As it can be observed, our method provides comparable, state-of-the-art, results with faster inference speed and smaller parameters. 
The qualitative results and comparison to previous methods are shown in Fig. \ref{fig:ablation1}, and more in Supplementary Materials.
As part of the inpainting process, the generated images are blended with agnostic images. 


\subsection{Ablation Study}

\begin{figure}[t!]
\centering
   \includegraphics[width=1.0\columnwidth]{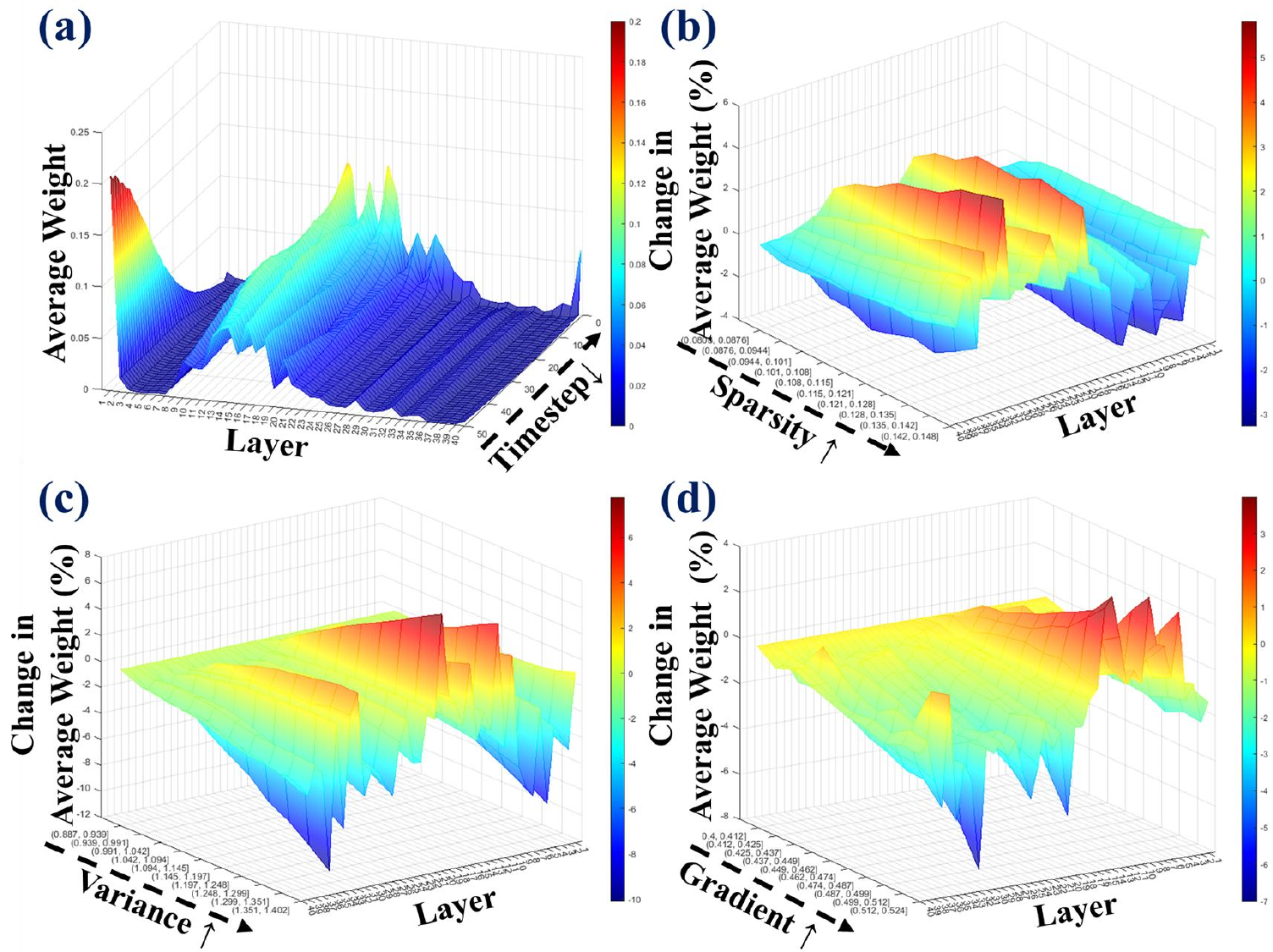}
   \caption{Visualization of the ITAFA's average weight distribution across ViT encoder layers of our final model.
   \textbf{(a):} use of hidden layers as denoising progresses. The dotted arrow refers to denoising direction. 
   \textbf{(b, c, d):} changes in the use of hidden layers as the sparsity, variance, and gradient magnitude of garment image changes. The dotted arrow refers to the increase in these image complexity components.}

\label{fig:itafa_ablation}
\end{figure}

\begin{figure*}[t!]
\centering
   \includegraphics[width=1.\linewidth]{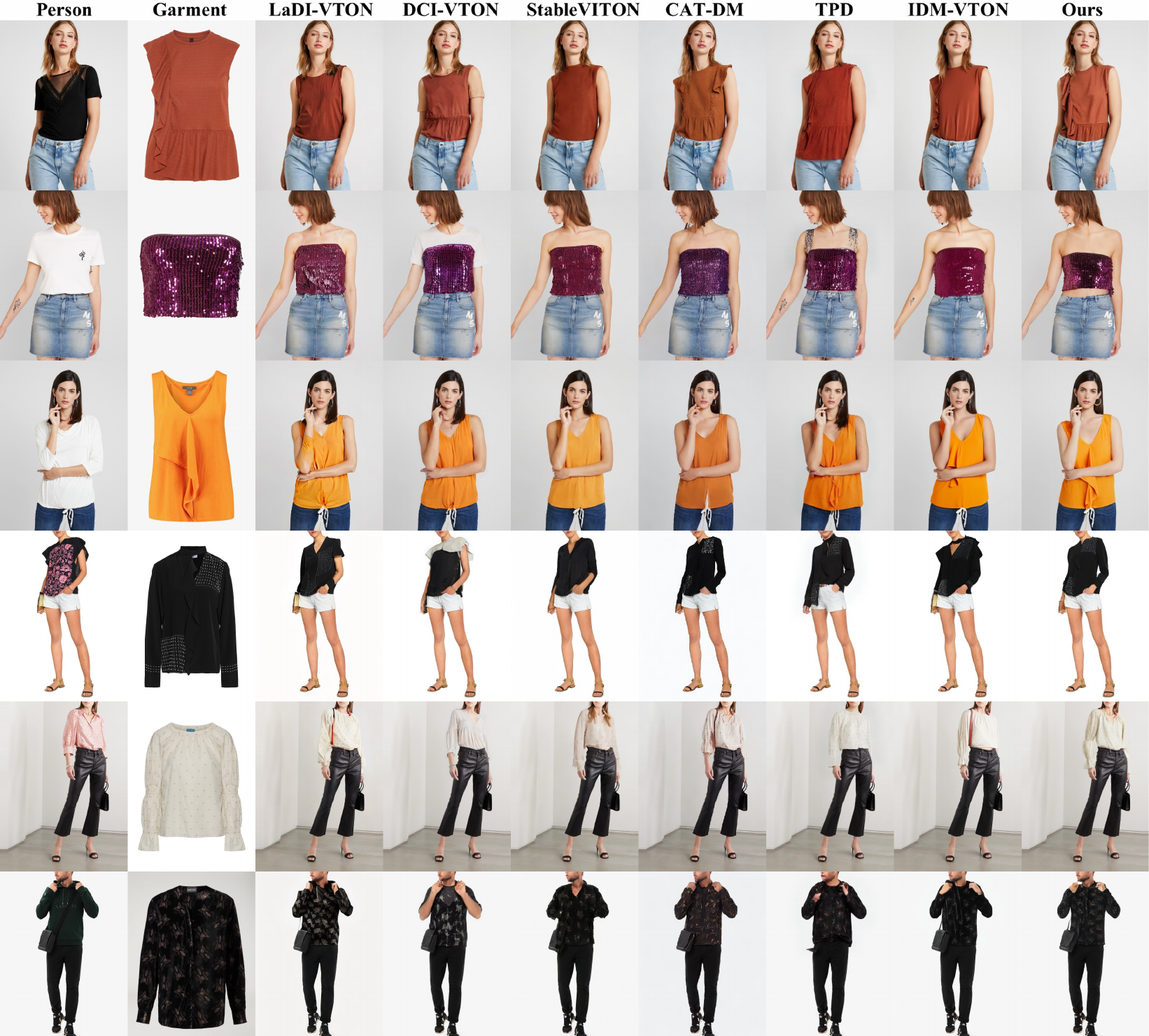}
   \caption{Qualitative comparison of ITA-MDT with prior methods on VITON-HD and DressCode Upper-body datasets. The first three rows show VITON-HD results; the next three show DressCode results.}

\label{fig:ablation1}
\end{figure*}

The qualitative results of our proposed ITA-MDT framework and effects of each components are available in the Supplementary Materials.
MDT-IVTON alone captures the global structure well, including garment shapes, colors, and human pose, but falls short in fidelity of garment details. The integration of ITAFA and SRE addresses this limitation significantly, with SRE shown as effective as a high-resolution image with twice the width compared to SRE integrated condition vector $c$.

Fig. \ref{fig:itafa_ablation} visualizes how ITAFA leverages features from different layers of the ViT image encoder, DINOv2, across the denoising process. 
In Fig. \ref{fig:itafa_ablation} (a), where it observes the use of layers with respect to timesteps, the middle layers dominate the feature aggregation for much of the diffusion; however, a distinct shift in layer usage occurs from early to later layers as the timestep progresses. 
In the early timestep when the input is largely noise, the encoder’s first layer is prominently utilized, and at the final timestep, the use of the last layer markedly increases. 
This observation supports our hypothesis that dynamically aggregating features from different layers as the diffusion progresses effectively optimizes feature utilization across denoising stages.
In Fig. \ref{fig:itafa_ablation} (b, c, d), we observe a clear pattern where specific layers are consistently favored depending on the intensity of particular image complexity (e.g., feature sparsity, variance, or gradient magnitude). This indicates that our model adaptively activates layers containing information most relevant to the type of visual complexity, highlighting the value and adaptability of our ITAFA.

\section{Conclusion}
\label{sec:conclusion}
This paper presented ITA-MDT, a framework for the IVTON task that successfully leverages MDT, along with the proposed ITAFA and SRE modules, to address the challenges of transformer-based denoising diffusion models in preserving fine-grained details efficiently. 
Analysis confirmed that ITAFA and SRE effectively function as intended, enabling ITA-MDT to achieve state-of-the-art performance in several metrics with reduced model parameters and faster inference speed compared to previous methods.

\section*{Acknowledgments}
This work was supported by Institute for Information \& communications Technology Planning \&
Evaluation (IITP) grant funded by the Korea government(MSIT) (No.RS-2021-II211381, Development of Causal AI through Video Understanding and Reinforcement Learning, and Its Applications to
Real Environments) and partly supported by Institute of Information \& communications Technology
Planning \& Evaluation (IITP) grant funded by the Korea government(MSIT) (No.RS-2022-II220184, Development and Study of AI Technologies to Inexpensively Conform to Evolving
Policy on Ethics)

{
    \small
    \bibliographystyle{ieeenat_fullname}
    \bibliography{main}
}

\clearpage
\setcounter{page}{1}
\maketitlesupplementary

\section{Masked Diffusion Transformer for Image-Based Virtual Try-On (MDT-IVTON)}
\label{sec:mdt-ivton_sup}
This section gives further explanation of \cref{sec:mdt-ivton}.

\subsection{Patchified Latent Formulation}
The Garment Agnostic Map, DensePose, and Garment Agnostic Mask images have initial shapes of \( A, P, M_X \in \mathbb{R}^{3, h, w} \), where \( h \) and \( w \) are the height and width, and 3 corresponds to the RGB channels. This image is encoded into the latent space by a VAE encoder, transforming them into \(\mathcal{E}(A), \mathcal{E}(P), \mathcal{E}(M_X) \in \mathbb{R}^{4, H, W} \), where \( H = h / 8 \) and \( W = w / 8 \). These are concatenated with the noised latent representation \( z_t \in \mathbb{R}^{4, H, W} \), resulting in a combined tensor \( \mathbb{R}^{16, H, W} \).

This combined tensor is patchified into a representation \( L \in \mathbb{R}^{p, D} \), with patch $p = \frac{H \cdot W}{\text{patch size}^2}$ and $D$ is the hidden layer embedding dimension.
Positional embeddings \( \in \mathbb{R}^{p, D} \) are added to \( L \), forming the final patchified latent representation. This representation undergoes denoising in the encoder and decoder blocks of MDT-IVTON. Within these blocks, \( L \) serves as the query in the cross-attention mechanism, interacting with the condition \( c \) to incorporate garment-specific information.

\subsection{Condition Formulation}
The garment image is processed through the Salient Region Extractor (SRE) and Image-Timestep Adaptive Feature Aggregator (ITAFA) to obtain the garment feature \( F_g \) and the salient region feature \( F_s \). These features have shapes \( F_g, F_s \in \mathbb{R}^{s, d} \), where \( s \) is the sequence length of the patch tokens from the image encoder and \( d \) is the embedding dimension of the image encoder.

Both \( F_g \) and \( F_s \) are projected to align with MDT-IVTON's embedding dimension \( D \), resulting in \( \mathbb{R}^{s, D} \). These are then concatenated along the sequence dimension to form \( \mathbb{R}^{2s, D} \). Time embedding \(T \in \mathbb{R}^{ D} \) is added to all sequences, formulating the final condition \( c \), which acts as the key and value in the cross-attention mechanism to guide denoising in MDT-IVTON.

\subsection{Denoising Objective}
The primary objective function minimizes the mean squared error (MSE) between the predicted noise and the actual noise in the noised latent \( z_t \) at each timestep \( t \), following the standard diffusion objective:
\begin{equation}
L_{\text{denoise}} = \mathcal{E}_{z_t, c, \epsilon \sim \mathcal{N}(0, I), t} \left[\lVert \epsilon - \epsilon_{\theta}(z_t, c, t)\rVert^2 \right],
\end{equation}
where \( z_t \in \mathbb{R}^{4, H, W} \) is the noised latent representation at timestep \( t \), \( c \in \mathbb{R}^{2s, D} \) is the condition, \( \epsilon \in \mathbb{R}^{4, H, W} \) is the Gaussian noise added during the forward diffusion process, and \( \epsilon_{\theta}(z_t, c, t) \in \mathbb{R}^{4, H, W} \) is the model's predicted noise.
This objective guides the model to learn to reverse the forward diffusion process.

\subsection{Mask Reconstruction Objective}
The Mask Reconstruction Objective operates on the masked latent representation \( L_m \), derived by applying a binary mask \( M_L \in \mathbb{R}^{p} \) to the patchified latent \( L \). The mask \( M_L \) indicates masked tokens with 0 and unmasked tokens with 1.

The Side-Interpolator reconstructs the masked tokens in \( L_m \) by leveraging the semantic information from the unmasked tokens as follows:
\begin{enumerate}
    \item The unmasked tokens \( L_u \in \mathbb{R}^{p_u, D} \), where \( p_u \) is the number of unmasked tokens, provide semantic context.
    \item The masked tokens \( L_m \in \mathbb{R}^{p_m, D} \), where \( p_m = p - p_u \), are reconstructed by interacting with \( L_u \) in the Side-Interpolator.
    \item The reconstruction leverages an attention mechanism:
    \begin{itemize}
        \item \( L_m \) serves as the query, representing the masked tokens requiring reconstruction.
        \item \( L_u \) serves as the key and value, encoding semantic and spatial context from the unmasked tokens.
        \item Attention weights computed between \( L_m \) (query) and \( L_u \) (key) determine how information from \( L_u \) (value) is used to reconstruct \( L_m \).
    \end{itemize}
    \item The output \( L'_m \in \mathbb{R}^{p_m, D} \) replaces the masked tokens in \( L_m \), forming a refined latent representation.
\end{enumerate}

The reconstruction loss is computed as:
\[
L_{\text{mask}} = \mathcal{E}_{y, c, t} \left[\lVert L'_m - L_m \rVert^2 \right],
\]
ensuring spatial coherence and semantic consistency in masked regions.

\subsection{Inpainting Objective}
\begin{figure}[h]
\centering
   \includegraphics[width=\columnwidth]{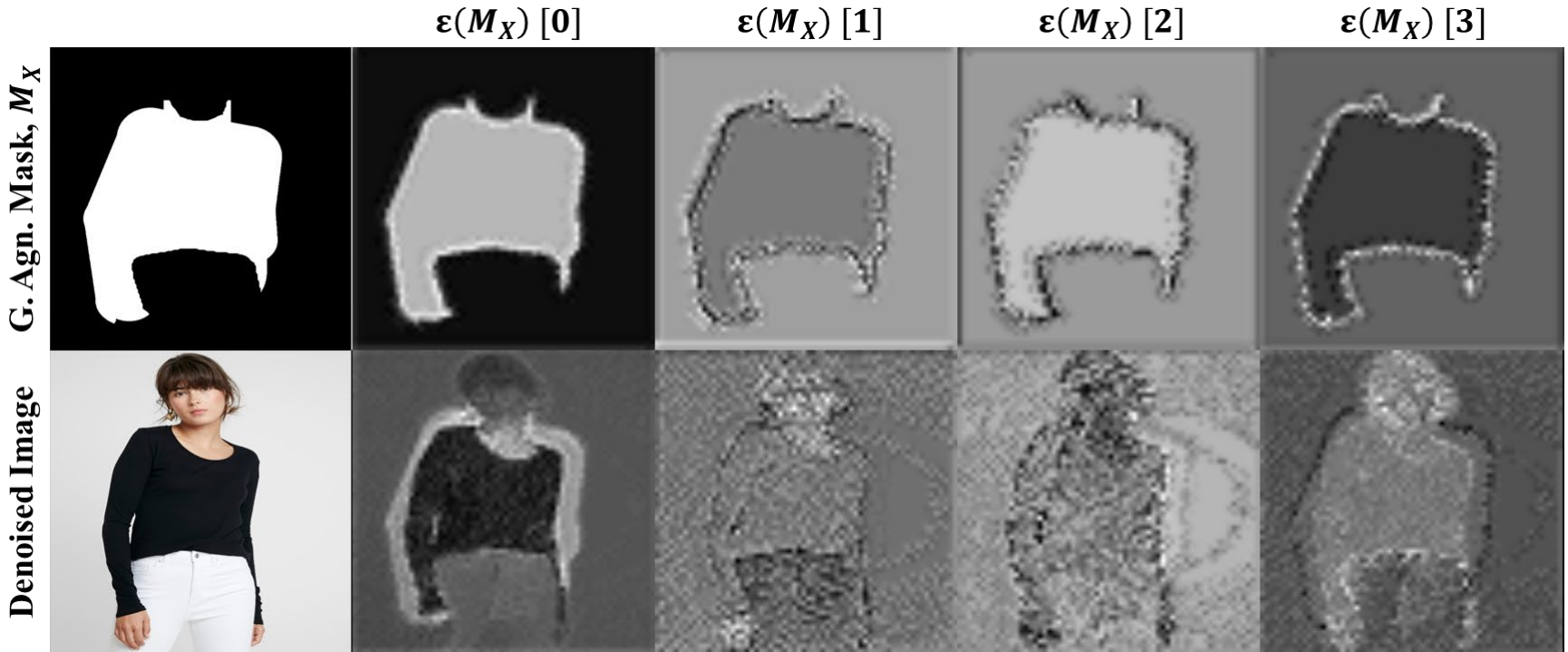}
   \caption{The top row visualizes the four normalized channels of the Garment Agnostic Mask in the latent space. The bottom row visualizes the corresponding four channels of the latent image, each with the mask from the matching channel applied.}
\label{fig:mdt_mask}
\end{figure}

The inpainting loss focuses on regions defined by the Garment Agnostic Mask \( M_X \). In the latent space, \( M_X \) is encoded as \( \mathcal{E}(M_X) \in \mathbb{R}^{4, H, W} \). Among its four channels, the first channel, \( \mathcal{E}(M_X)_0 \in \mathbb{R}^{H, W} \), closely resembles the binary mask and is used for loss computation, which is visualized in Fig. \ref{fig:mdt_mask}

This channel is normalized to ensure numerical stability, and the inpainting loss is calculated as:
\[
L_{\text{inpaint}} = \mathcal{E}_{y, c, t} \left[\lVert \mathcal{E}(M_X)_0 \cdot (y_t - y_{t-1}) \rVert^2 \right],
\]
where \( y_t, y_{t-1} \in \mathbb{R}^{4, H, W} \) are latent representations at consecutive timesteps. This loss emphasizes garment-specific details in masked regions.

\subsection{Overall Loss Function}
The overall loss integrates denoising, mask reconstruction, and inpainting objectives:
\[
L_{\text{total}} = L_{\text{denoise}} + L_{\text{mask}} + L_{\text{inpaint}}.
\]
This formulation balances global structure, spatial coherence, and fidelity of garment-specific details, ensuring high-quality virtual try-on results.

\section{Image-Time Adaptive Feature Aggregator (ITAFA)}
\label{sec:itafa_sup}
This section elaborates on the ITAFA of \cref{sec:sre} in detail. 
\paragraph{Feature Complexity Components.}
Complexity components for the input feature tensor $f$ are calculated as below. 
\subsection{Feature Sparsity} $S$ quantifies the proportion of near-zero activations in the feature embeddings, providing insight into the sparsity of structural activations within $f$. Given a threshold $\delta$, sparsity is defined as:
        \begin{equation}
        S = \frac{1}{H \times s \times d} \sum_{i=1}^{H} \sum_{j=1}^{s} \sum_{k=1}^{d} \mathbb{I}(|f_{ijk}| < \delta),
        \end{equation}
where $H$ is the number of hidden layers of the image encoder, $s$ is the sequence length of the patch tokens, $d$ is the embedding dimension, and $\mathbb{I}$ is the indicator function that equals 1 when $|f_{ijk}| < \delta$ and 0 otherwise. This function allows for counting the proportion of elements in $f$ that are below the threshold (i.e., near zero) providing a measure of sparsity.

\subsection{Feature Variance} $V$ reflects the variability across activations, capturing structural complexity and richness of detail in $f$:
        \begin{equation}
        V = \frac{1}{H \times s \times d} \sum_{i=1}^{H} \sum_{j=1}^{s} \sum_{k=1}^{d} (f_{ijk} - \overline{f})^2
        \end{equation}
    where $\overline{f}$ is the mean activation across all embeddings.

\subsection{Gradient Magnitude} $G$ measures local variations in feature embeddings by computing spatial gradients along the sequence and embedding dimensions. This component captures texture and fine details, calculated as:
\begin{equation}
    \Delta f_{i,j,k} = \sqrt{(f_{i,j+1,k} - f_{i,j,k})^2 + (f_{i,j,k+1} - f_{i,j,k})^2}
\end{equation}
\begin{equation}
    G = \frac{1}{H \times (s-1) \times (d-1)} \sum_{i=1}^{H} \sum_{j=1}^{s-1} \sum_{k=1}^{d-1} \Delta f_{i,j,k}
\end{equation}

\( \Delta f_{i,j,k} \) represents the gradient magnitude at each index \((i, j, k)\) in the feature embedding tensor. \( G \) then averages the \( \Delta f \) values over the entire feature embedding tensor to capture the overall texture complexity.

The combined complexity score vector, $[S, V, G] \in \mathbb{R}^3$, captures the structural and textural complexity of $f$.

\section{Salient Region Extractor (SRE)}
\label{sec:sre_sup}
This section describes the SRE algorithm in detail, elaborating on \cref{sec:sre}.

\subsection{Entropy Map Computation}
The entropy map \( X_e \) provides a measure of the information content for each pixel in the grayscaled version of the input garment image, \( \mathcal{X}_{\text{gray}} \). The Shannon Entropy \cite{shannon1948mathematical} is utilized to capture local texture complexity and information density.

\paragraph{Local Neighborhood Definition.}
For each pixel \( \mathcal{X}_{\text{gray}}(i, j) \), we consider a local neighborhood of size \( 5 \times 5 \), denoted as \( N_{i,j} \), centered at pixel \((i, j)\):
\begin{equation}
    N_{i,j} = \{ \mathcal{X}_{\text{gray}}(m, n) \mid i-2 \le m \le i+2, \; j-2 \le n \le j+2 \},
\end{equation}
where the neighborhood is truncated near the image borders to fit within the image dimensions.

\paragraph{Probability Distribution of Intensities.}
In each neighborhood \( N_{i,j} \), we count the frequency of each pixel intensity value from 0 to 255, and calculate the probability \( p_k \) of each intensity value \( k \) as:
\begin{equation}
    p_k = \frac{1}{|N_{i,j}|} \sum_{(m, n) \in N_{i,j}} \mathbb{I}(\mathcal{X}_{\text{gray}}(m, n) = k),
\end{equation}
where \( |N_{i,j}| = 25 \) is the number of pixels in the neighborhood, and $\mathbb{I}(\cdot)$ is the indicator function that returns 1 if the condition is true and 0 otherwise.

\paragraph{Shannon Entropy Calculation.}
The Shannon entropy \( H_{i,j} \) for the neighborhood \( N_{i,j} \) is calculated as:
\begin{equation}
    H_{i,j} = -\sum_{k=0}^{255} p_k \log_2(p_k),
\end{equation}
where \( p_k \) represents the probability of intensity \( k \) within the neighborhood. If \( p_k = 0 \), the corresponding term is considered zero, as \( p_k \log_2(p_k) = 0 \) for \( p_k = 0 \).

\paragraph{Constructing the Entropy Map.}
The entropy map \( X_e \) is constructed by assigning the computed entropy value \( H_{i,j} \) to each pixel \((i, j)\) in the image:
\begin{equation}
    X_e(i, j) = H_{i,j},
\end{equation}
resulting in an entropy map \( X_e \in \mathbb{R}^{H \times W} \) that provides a grayscale representation of the information content for each pixel.

\paragraph{Interpretation of the Entropy Map.}
The resulting entropy map \( X_e \) reflects the complexity of each region in the image:
\begin{itemize}
    \item \textbf{High Entropy}: Regions with higher entropy indicate greater variability in pixel intensities, suggesting areas with rich textures, edges, or details.
    \item \textbf{Low Entropy}: Regions with lower entropy represent uniform areas with little variation.
\end{itemize}
This process effectively highlights the most informative areas of the input image.

\subsection{High-Entropy Region Identification}
To isolate regions of interest, a binary mask $X_m$ is generated by thresholding the entropy map with a pre-defined entropy threshold $E$:
\begin{equation}
X_{\text{m}}(i,j) = \begin{cases} 
      1 & \text{if } X_e(i,j) > E, \\
      0 & \text{otherwise}.
   \end{cases}
\end{equation}
The entropy threshold $E$ is empirically set to 0.8.
If no high-entropy regions are detected, adaptive thresholding is applied, gradually lowering the threshold until a region is found or reaches a minimum value. If this adjustment fails.

\subsection{Entropy Centroid Localization}
The centroid $(x_c, y_c)$ of the high-entropy region is computed as the center of mass:
\begin{equation}
(x_c, y_c) = \frac{\sum_{i,j} X_m(i,j) \cdot (i,j)}{\sum_{i,j} X_m(i,j)}.
\end{equation}
If no high-entropy regions are found, the fallback behavior sets the centroid to the image center, ensuring robustness in cases of low entropy.

\subsection{Circular Region Expansion}
The region around the centroid is initially bounded by a square of width and height $l_{min}$, set to 224, centered at $(x_c, y_c)$. 
$l_{min}$ defines the minimum height and width of the Salient Region to prevent overly small regions and extreme aspect ratios.
This bounding box expands outward in a circular pattern (i.e., up, right, down, left), repeating this sequence until no further expansion is needed. For each direction, the algorithm checks whether the newly added edge pixels contain more high-entropy pixels than a given threshold to determine if expansion should continue.

\subsection{Region Extraction}
After the Circular Region Expansion, the bounding box is adjusted to match the aspect ratio of the original image to minimize distortion. The adjustment involves expanding either the height or width, depending on the current bounding box's aspect ratio compared to the original \( \mathcal{X}_{\text{gray}} \). The final Salient Region $X_s$ extracted maintains the aspect ratio, preserving visual consistency. 
By preserving the original aspect ratio, the model can effectively perceive the salient region in the context of the full garment, minimizing potential spatial confusion.
The extracted region is then resized to 224×224 to be processed as a diffusion condition.

\section{Denoising with Classifier-Free Guidance}

The denoising process employs Classifier-Free Guidance (CFG) \cite{ho2022classifier} to dynamically balance unconditional and conditional noise predictions during each timestep \( t \) of the diffusion process. 

\subsection{DDIM Sampling}

The iterative denoising process is implemented using a DDIM sampling loop, which refines the noisy latent representation \( z_t \in \mathbb{R}^{4, H, W} \) over a predefined number of diffusion steps \( \gamma_{\text{steps}} \). At each timestep, the model predicts the noise \( \epsilon_t \) to compute the latent representation for the next timestep \( z_{t-1} \):
\begin{equation}
z_{t-1} = \text{DDIM}(\epsilon_t, z_t, t, \gamma_{\text{steps}}),
\end{equation}
where the process continues until \( z_0 \), clean latent representation. The sampling loop iteratively combines conditional and unconditional noise predictions using the guidance mechanism described below.

\subsection{Unconditional and Conditional Predictions}

At each timestep \( t \), the noisy latent representation \( z_t \) is processed by the denoising model \( \epsilon_{\theta} \) to produce two noise predictions:
\begin{align}
\epsilon_{\text{uncond}} &= \epsilon_{\theta}(z_t, t, \emptyset), \\
\epsilon_{\text{cond}} &= \epsilon_{\theta}(z_t, t, c),
\end{align}
where \( \epsilon_{\text{uncond}} \in \mathbb{R}^{4, H, W} \) is the unconditional noise prediction generated without the condition $c$, and \( \epsilon_{\text{cond}} \in \mathbb{R}^{4, H, W} \) is the noise prediction with condition \( c \) informed.

\subsection{Dynamic CFG Scaling}

The classifier-free guidance scale \( \alpha_{\text{cfg}} \) determines the strength of the conditional guidance, which is dynamically adjusted at each timestep using a cosine-based power scaling function:
\begin{equation}
\delta_{\text{scale}} = \frac{1 - \cos\left(\left(1 - \frac{t}{\gamma_{\text{steps}}}\right)^{\beta_{\text{scale}}} \cdot \pi\right)}{2},
\end{equation}
where \( \gamma_{\text{steps}} \) is the total number of diffusion steps, and \( \beta_{\text{scale}} \) is the power scaling factor. The intermediate scale factor \( \delta_{\text{scale}} \in [0, 1] \) ensures a smooth adjustment of guidance strength, starting with weaker conditional emphasis in earlier steps and gradually increasing its influence.

The effective guidance scale \( \alpha_{\text{eff}} \) for timestep \( t \) is computed as:
\begin{equation}
\alpha_{\text{eff}} = 1 + (\alpha_{\text{cfg}} - 1) \cdot \delta_{\text{scale}}.
\end{equation}

\subsection{Guided Noise Prediction}

The final noise prediction \( \epsilon_t \) is a weighted combination of the unconditional and conditional predictions:
\begin{equation}
\epsilon_t = \epsilon_{\text{uncond}} + \alpha_{\text{eff}} \cdot (\epsilon_{\text{cond}} - \epsilon_{\text{uncond}}),
\end{equation}
where \( \epsilon_t \in \mathbb{R}^{4, H, W} \). 

\section{Implementation Details}
\label{sec:imp_details_sup} 
This section provides further details on the implementation of the proposed model, as outlined in \cref{sec:imp_details}. The foundational architecture for MDT-IVTON is derived from the MDTv2 XL model, configured with a depth of 28 layers, including 4 decoding layers, a hidden size of 1152, and 16 attention heads per layer. We utilize RGB images of size \( 512 \times 512 \times 3 \) as both the input reference images and the generated output results. To ensure a fair comparison with prior works, the Variational Autoencoder (VAE) from Stable Diffusion XL \citep{podell2023sdxl} is employed, encoding images into a latent representation \( z \) with dimensions \( 64 \times 64 \times 4 \). 

During training, we use 1000 diffusion steps, whereas 30 steps are used for the generation results reported in Table \ref{tab:qual_table}. The mask ratio in the MDT training scheme is set to 0.3. For optimization, we use an initial learning rate of \( 1 \times 10^{-4} \) with a batch size of 6. Training stability is enhanced through the use of an Exponential Moving Average (EMA) with a rate of 0.9999. All other hyperparameters, including the optimizer and learning rate scheduler, follow the configurations used in DiT \citep{peebles2023scalable}, ensuring consistency with existing diffusion-based transformer approaches. For inference, the model is evaluated using 30 sampling steps \( \gamma_{\text{steps}} \). The Classifier-Free Guidance scale \( \alpha_{\text{cfg}} \) is set to 2.0, and the power scaling factor \( \beta_{\text{scale}} \) is set to 1.0. We report the performance of the model trained for 2 million steps.

\section{Dataset and Evaluation Metrics}
\label{sec:data_metric_details_sup}
This section gives details on datasets used and the evaluation metrics, mentioned in  \cref{sec:data_metric_details}.
The VITON-HD dataset comprises 13,679 images of human models paired with upper garment images, with person images provided at a resolution of \(1024 \times 768\). This dataset is widely used for virtual try-on tasks and features relatively simple poses, where subjects stand in straightforward, static positions with solid-colored backgrounds. The simplicity of backgrounds and poses makes VITON-HD an ideal testbed for evaluating the model's performance in basic virtual try-on scenarios.

The DressCode dataset offers a more diverse collection, containing over 50,000 high-resolution images (\(1024 \times 768\)) across three categories: upper-body garments, lower-body garments, and dresses. Specifically, it includes 17,650 images of upper-body garments, 17,650 images of lower-body garments, and 17,650 images of dresses. Similar to VITON-HD, DressCode features consistent, simpler poses against plain backgrounds. However, it includes a more diverse set of garment styles, offering additional challenges in fitting and transferring intricate patterns, logos, and textures accurately during virtual try-on tasks. Both datasets serve as benchmarks, with VITON-HD focusing on basic pose and background handling, and DressCode testing the model's ability to preserve detailed garment features across various clothing types.

To evaluate performance, we employ several widely-used metrics in virtual try-on research. LPIPS (Learned Perceptual Image Patch Similarity) measures perceptual similarity by comparing deep features from neural networks, with lower LPIPS scores indicating greater perceptual closeness to ground truth. SSIM (Structural Similarity Index) evaluates the structural integrity of generated images by quantifying similarity in luminance, contrast, and structure; higher SSIM values indicate better preservation of the original structure. 
FID (Fréchet Inception Distance) assesses quality and diversity by comparing the feature distributions of generated and real images, with lower FID values denoting closer alignment to real image distributions. 
We report both paired and unpaired FID results. While FID is commonly used to assess unpaired results in VTON task, paired FID is also informative as it directly compares generated images with their corresponding ground-truth images, which does not exist for unpaired generation.

\section{Ablation Study}
\subsection{Analysis on ITAFA}
For the learnable parameter $\alpha$ of Eq. \eqref{eq_alpha} which controls the balance between timestep information and image complexity in the aggregation process, the final value of our final model is $0.655$. This indicates that the model emphasizes timestep information, while still incorporating image complexity.

The image complexity distribution of the garments in VITON-HD dataset and DressCode dataset are organized in Table \ref{tab:itafa_ablation}. 

\begin{table}[ht]
    \centering
    \begin{tabular}{lcccccc}
        \toprule
        \textbf{Data} & \multicolumn{2}{c}{\textbf{Sparsity}} & \multicolumn{2}{c}{\textbf{Variance}} & \multicolumn{2}{c}{\textbf{Gradient}} \\
        & \textbf{Avg.} & \textbf{Std.} & \textbf{Avg.} & \textbf{Std.} & \textbf{Avg.} & \textbf{Std.} \\
        \midrule
        V-HD & 0.125 & 0.011 & 1.139 & 0.077 & 0.462 & 0.013 \\
        DC-U & 0.127 & 0.010 & 1.156 & 0.081 & 0.459 & 0.015 \\
        DC-L & 0.128 & 0.008 & 1.161 & 0.065 & 0.450 & 0.014 \\
        DC-D & 0.129 & 0.008 & 1.157 & 0.057 & 0.456 & 0.015 \\
        \bottomrule
    \end{tabular}
    \caption{The V-HD, DC-U, DC-L, and DC-D denote the VITON-HD, DressCode Upper-body subset, Lower-body subset, and Dresses subset, respectively. \textbf{Avg.} and \textbf{Std.} denote the average and standard deviation of the values, respectively.}
    \label{tab:itafa_ablation}
\end{table}

The results highlight subtle differences between the VITON-HD and DressCode datasets, particularly for upper-body garments. DressCode garments exhibit slightly higher sparsity and variance compared to those in VITON-HD, suggesting that individual garments in DressCode may contain patterns with relatively greater complexity, such as logos or prints, which contribute to increased pixel intensity variations.
In contrast, the lower average gradient magnitude in DressCode samples indicates that these patterns often have smoother transitions or softer boundaries, likely due to similar colors between garments and their prints or repetitive designs with subtle changes. Meanwhile, the relatively higher gradient values in VITON-HD garments suggest that they may include simpler, more distinct patterns, such as large logos with sharp edges and contrasting colors.

Although the differences exist, they are subtle. 
The similar garment complexity across the datasets explains the small gap in FID scores between the VITON-HD and DressCode Upper-body datasets in Table \ref{tab:qual_table}. The performance of our model on DressCode dataset, including Lower-body and Dresses, are shown in Table \ref{tab:qual_dc}.

\begin{table}[h]
    \centering
    \begin{tabular}{l|ccc}
        \toprule
        Subset & LPIPS↓ & SSIM↑ & FID(p./unp.)↓ \\
        \midrule
        Upper-body & 0.034 & 0.951 & 5.412/10.069 \\
        Lower-body & 0.052 & 0.931 & 6.109/12.335 \\
        Dresses & 0.080 & 0.883 & 6.957/10.662 \\
        \bottomrule
    \end{tabular}
    \caption{Performance of our ITA-MDT on three subsets of DressCode dataset. \textbf{p.} and \textbf{unp.} denotes paired and unpaired generation evaluation, respectively.
    }
    \label{tab:qual_dc}
\end{table}

\subsection{Analysis on SRE}
\begin{figure*}[th]
\centering
   \includegraphics[width=1.0\textwidth]{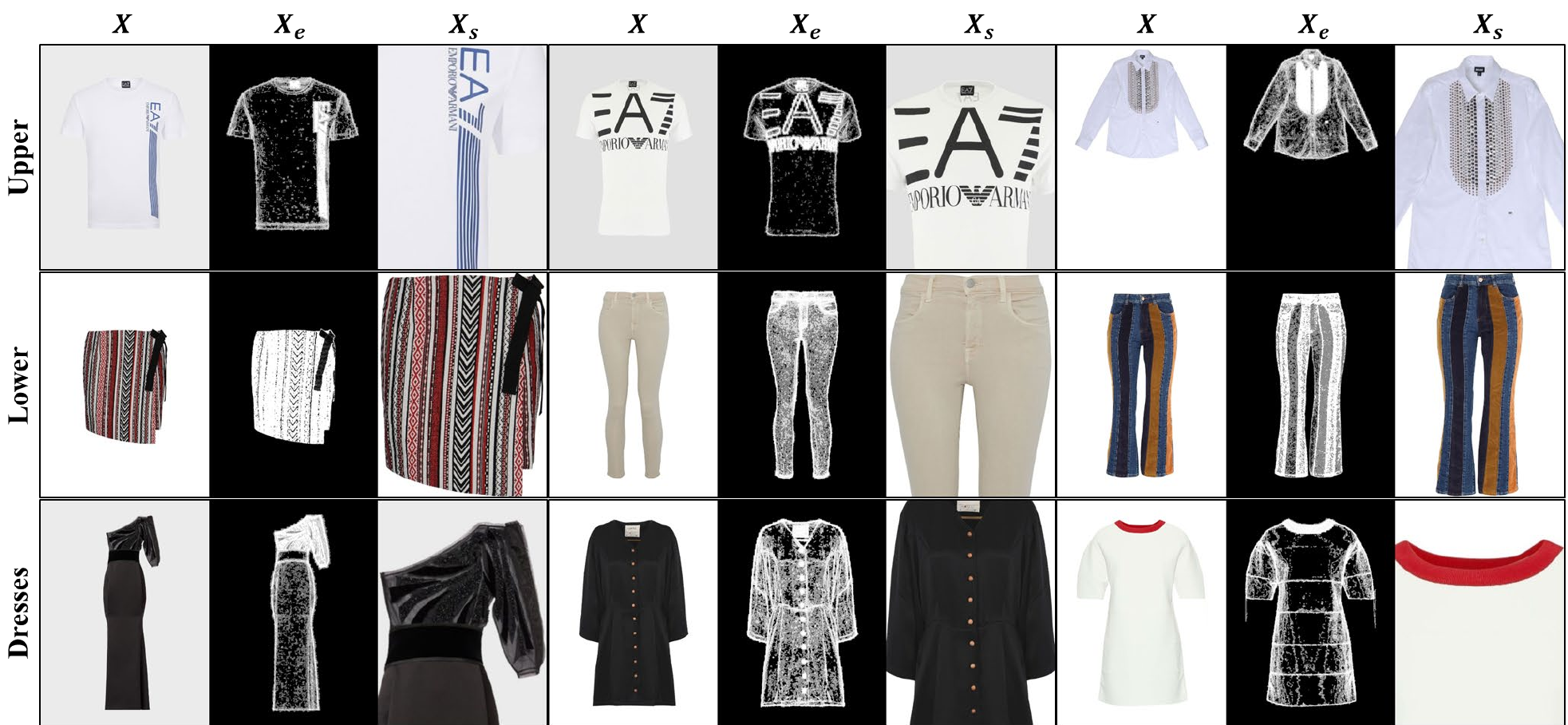}
   \caption{Garment image $X$ with its Entropy Map $X_e$ and its Salient Region $X_s$ extracted with our Salient Region Extractor. Images are from the upper-body, lower-body, and dresses subset of the DressCode dataset.}
\label{fig:sre_samples}
\end{figure*}

\begin{figure}[h]
  \centering
  \includegraphics[width=0.8\linewidth]{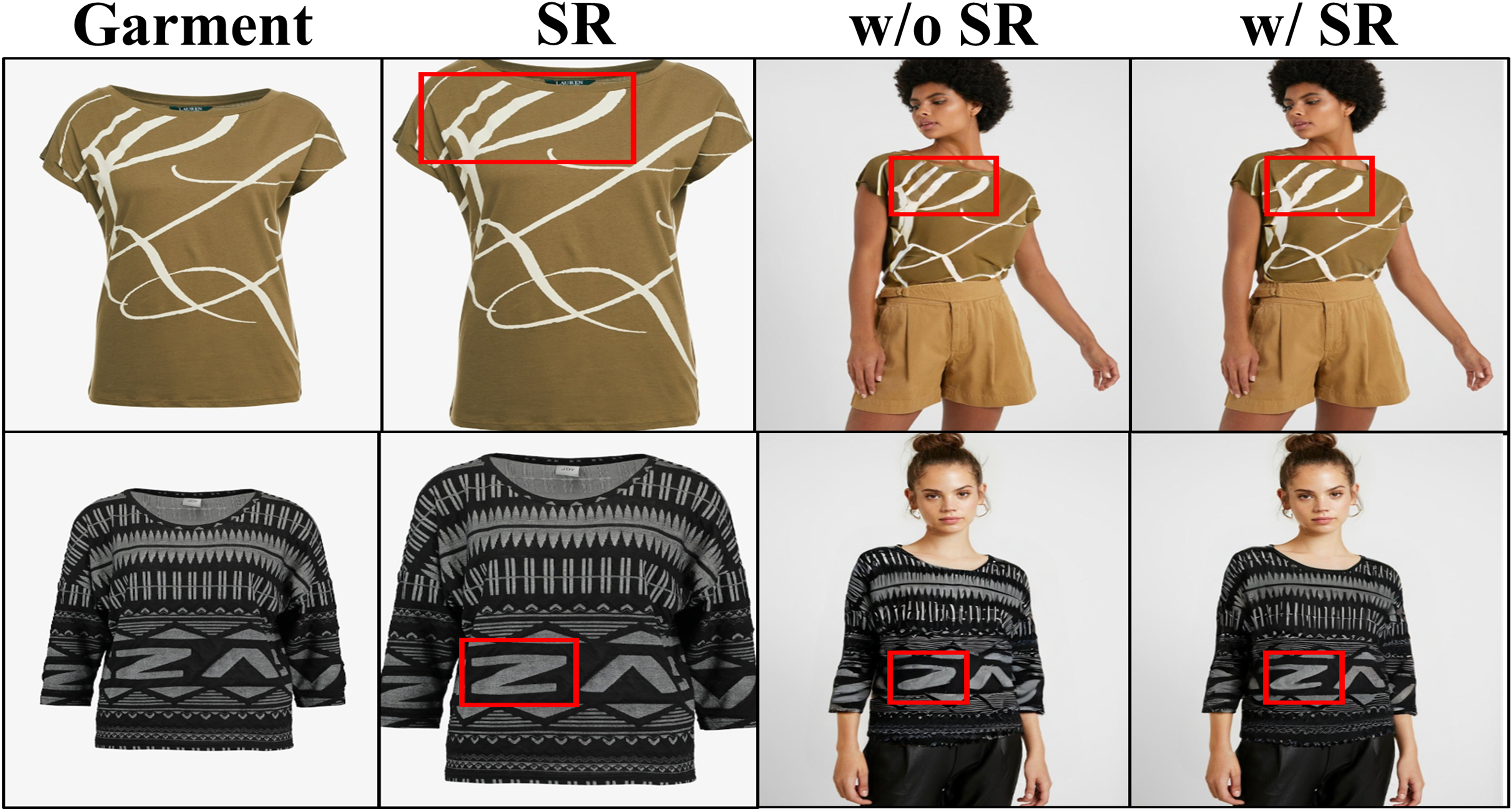}
   \caption{Results with and without Salient Region (SR).}
   \label{fig:rebuttal_clusters}
\end{figure}

\begin{figure}[h]
  \centering
  \includegraphics[width=\linewidth]{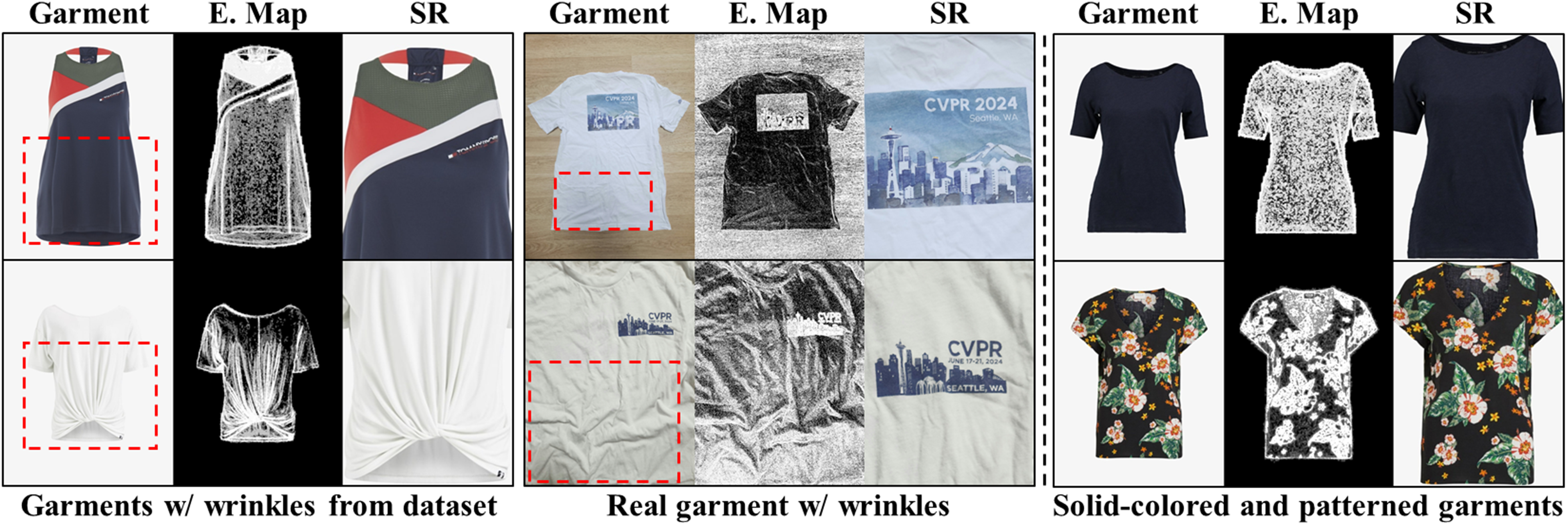}
   \caption{Salient Region Extractor (SRE) outputs: E. Map and SR refer to Entropy Map and Salient Region, respectively. Red box indicates wrinkled regions.}
   \label{fig:rebuttal_sre}
\end{figure}

Figure~\ref{fig:sre_samples} illustrates examples of the Entropy Map $X_e$ and the corresponding extracted Salient Region $X_s$ from a given garment image $X$. 
When no dominant high-entropy cluster is detected, such as solid-colored and uniformly patterned garments, SRE tends to extract a broader region or even the entire garment, as shown in  Figure~\ref{fig:rebuttal_sre} (right). 
While this may seem less selective, it remains beneficial by reducing background region, capturing the full garment with higher-resolution information that enriches local detail representation. 
Examples of such cases are shown in Figure~\ref{fig:rebuttal_clusters}.

A potential concern is whether high-entropy elements such as wrinkles (possibly irrelevant to the target garment) might be mistakenly emphasized. 
However, in the virtual try-on setting, garments are photographed under controlled conditions, where such elements are likely part of the intended design. 
As a result, our entropy-based method remains robust for this task.
Figure~\ref{fig:rebuttal_sre} (left) shows that SRE performs well even on garments with wrinkles, using both dataset and real-world examples.

To accelerate training and evaluation, Salient Regions were preprocessed in advance. The SRE process averaged about 1.563 seconds per image in our experimental environment.

\subsection{Analysis on Mask Reconstruction Objective}
\begin{figure}[h!]
\centering
   \includegraphics[width=\columnwidth]{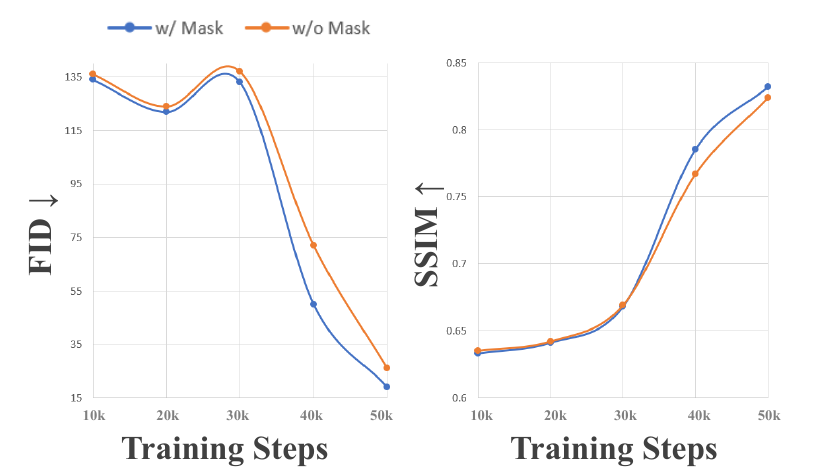}
   \caption{Comparison of training efficiency between ITA-MDT with and without the mask reconstruction objective with side-interpolator of Masked Diffusion Transformer(MDT). Evaluated with VITON-HD paired.}
\label{fig:mask_effect}
\end{figure}

The mask reconstruction objective of the Masked Diffusion Transformer (MDT) with its Side-Interpolator module is the key component that transforms a Diffusion Transformer (DiT) into an MDT.
The impact of the mask reconstruction objective on training efficiency and performance is illustrated in Figure \ref{fig:mask_effect}. The figure compares the early training progression of our ITA-MDT with and without the mask reconstruction objective. The mask reconstruction objective produces a steeper learning curve, indicating faster convergence.

\subsection{Analysis on Sampling Steps}
The number of sampling steps \( \gamma_{\text{steps}} \) used during inference directly influences the trade-off between inference speed and the quality of generated results. Table \ref{tab:steps_tradeoff} highlights this trade-off.
Higher \( \gamma_{\text{steps}} \) leads to improvement in FID but slight degradation in SSIM and LPIPS. 
This discrepancy may have arisen because higher sampling steps refine fine-grained textures, which improve perceptual quality captured by FID, but may slightly alter pixel-level structural consistency, affecting SSIM and LPIPS. Additionally, the longer sampling trajectory may have introduced small deviations in structure as the latent representation evolves.

While we balance these factors by using \( \gamma_{\text{steps}} = 30 \), a reduction of \( \gamma_{\text{steps}} \) can be considered for accelerated use cases where inference speed is prioritized over marginal improvements in perceptual quality.

\begin{table}[h]
\centering
    \begin{tabular}{c|c|c|c|c}
    \hline
    \( \gamma_{\text{steps}} \) & Inf. Time (s)  & LPIPS↓ & SSIM↑ & FID↓ \\
    \hline
    20 & \textbf{3.207} & 0.084 & \textbf{0.888} & 5.799 \\
    25 & 3.912 & 0.084 & \textbf{0.888} & 5.594 \\
    30 & 4.606 & \textbf{0.083} & 0.885 & 5.462 \\
    35 & 5.284 & \textbf{0.083} & 0.885 & 5.355 \\
    40 & 5.951 & \textbf{0.083} & 0.885 & 5.322 \\
    45 & 6.938 & \textbf{0.083} & 0.884 & 5.347 \\
    50 & 7.427 & 0.084 & 0.884 & \textbf{5.293} \\
    \hline
    \end{tabular}
    \caption{Trade-off between sampling steps (\( \gamma_{\text{steps}} \)), Inference Time (Inf. Time), and image quality metrics on VITON-HD paired evaluation. Note that the optimal classifier-free guidance scale \( \alpha_{\text{cfg}} \) and power scaling factor \( \beta_{\text{scale}} \) of our model were determined using \( \gamma_{\text{steps}} \) of 30.}
    \label{tab:steps_tradeoff}
\end{table}

\newpage

\section{Qualitative Results and Comparison}
\label{sec:qual_sup}
We provide a detailed overview of the qualitative results of our ITA-MDT framework and its comparison to previous methods, highlighting its superior fidelity in capturing the texture and color of the garments.

\begin{figure*}[t]
\centering
   \includegraphics[width=1.0\textwidth]{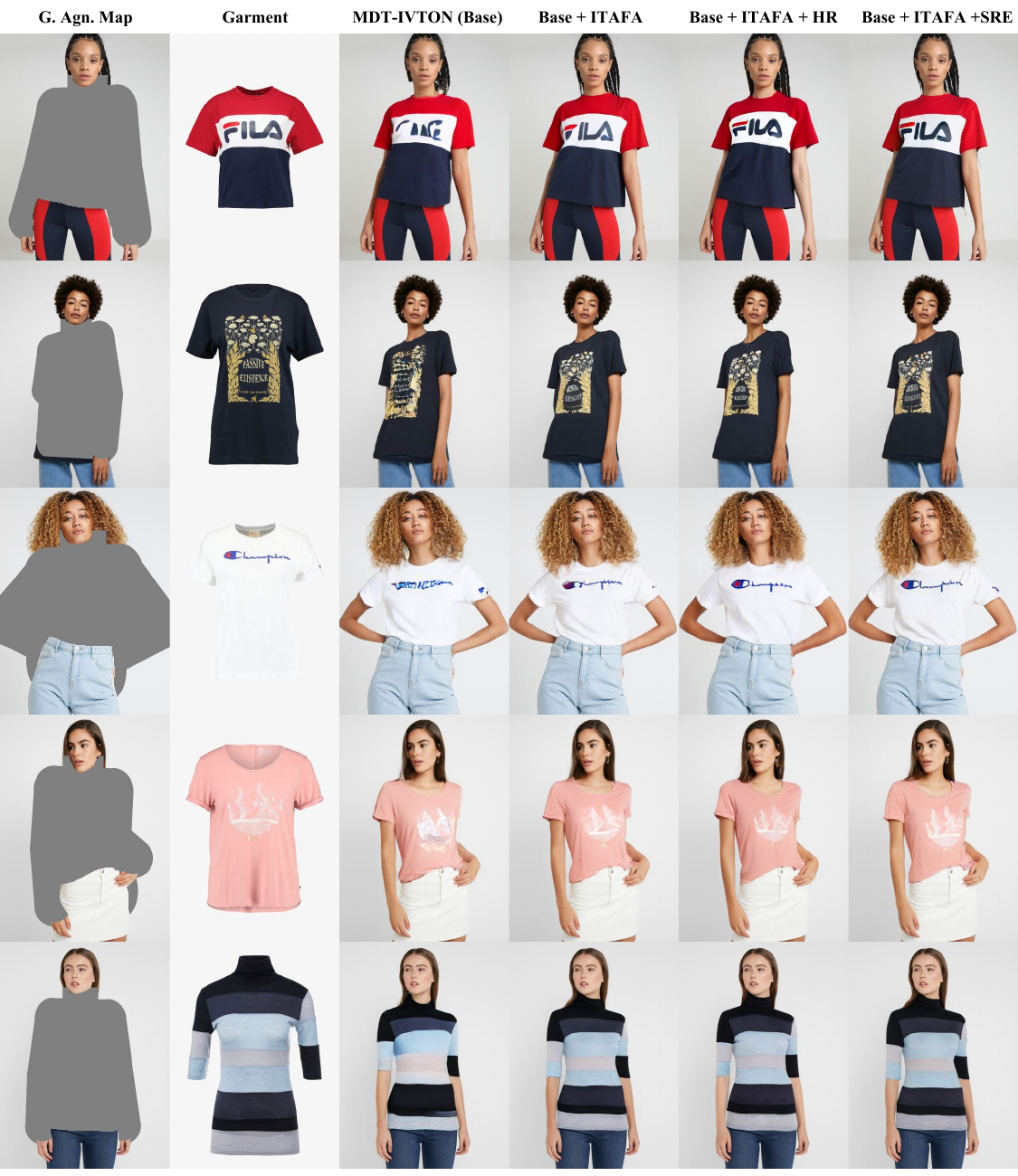}
   \caption{Qualitative comparison of the effect of each component of the ITA-MDT framework on VITON-HD. HR refers to the use of single High-resolution $(448 \times 448 \times 3)$ garment image to formulate condition vector $c$.}
\label{fig:qual1}
\end{figure*}

\begin{figure*}[t]
\centering
   \includegraphics[width=0.8\textwidth]{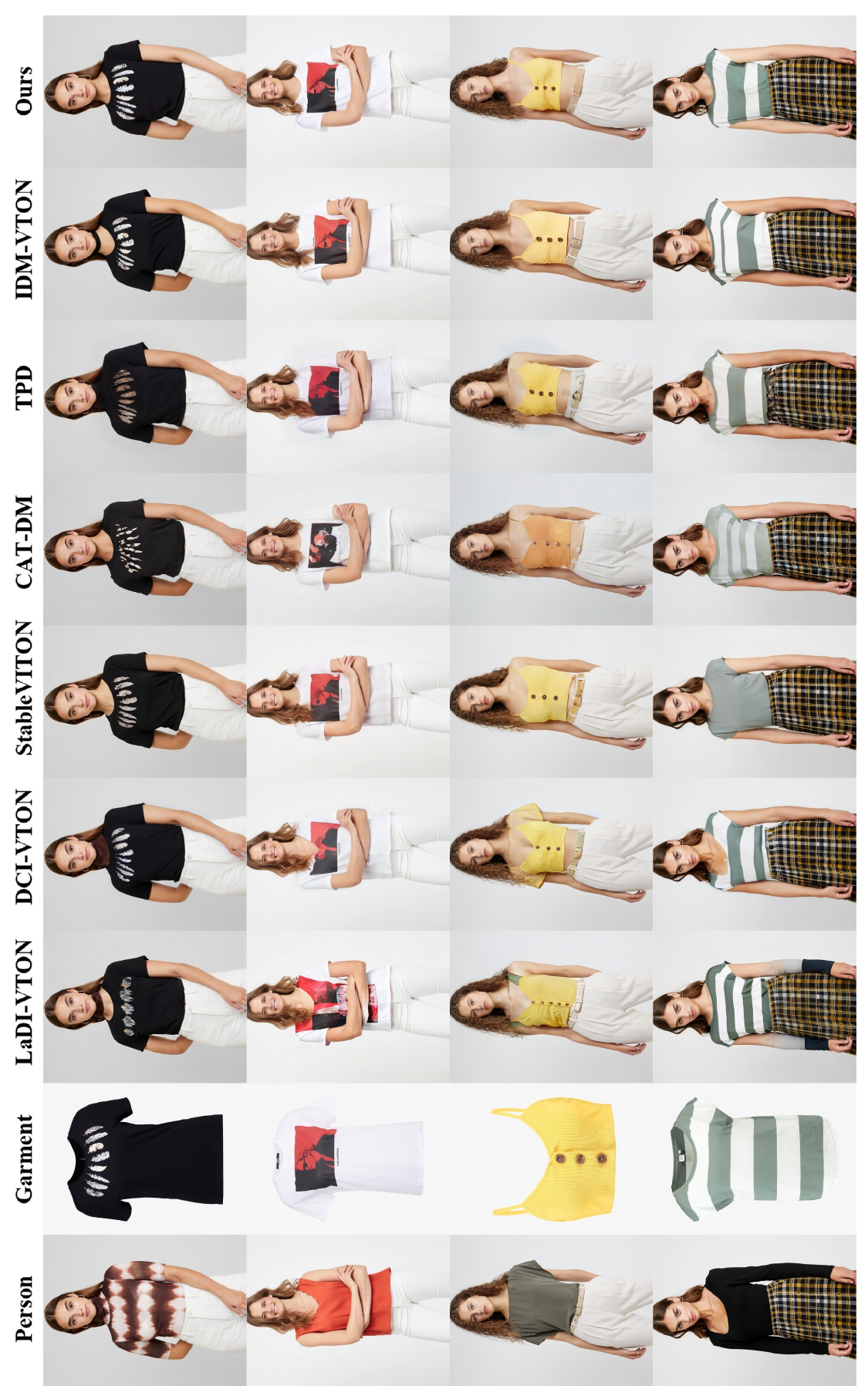}
   \caption{Qualitative comparison between our ITA-MDT and previous methods on the VITON-HD.}
\label{fig:qual2}
\end{figure*}

\begin{figure*}[t]
\centering
   \includegraphics[width=0.8\textwidth]{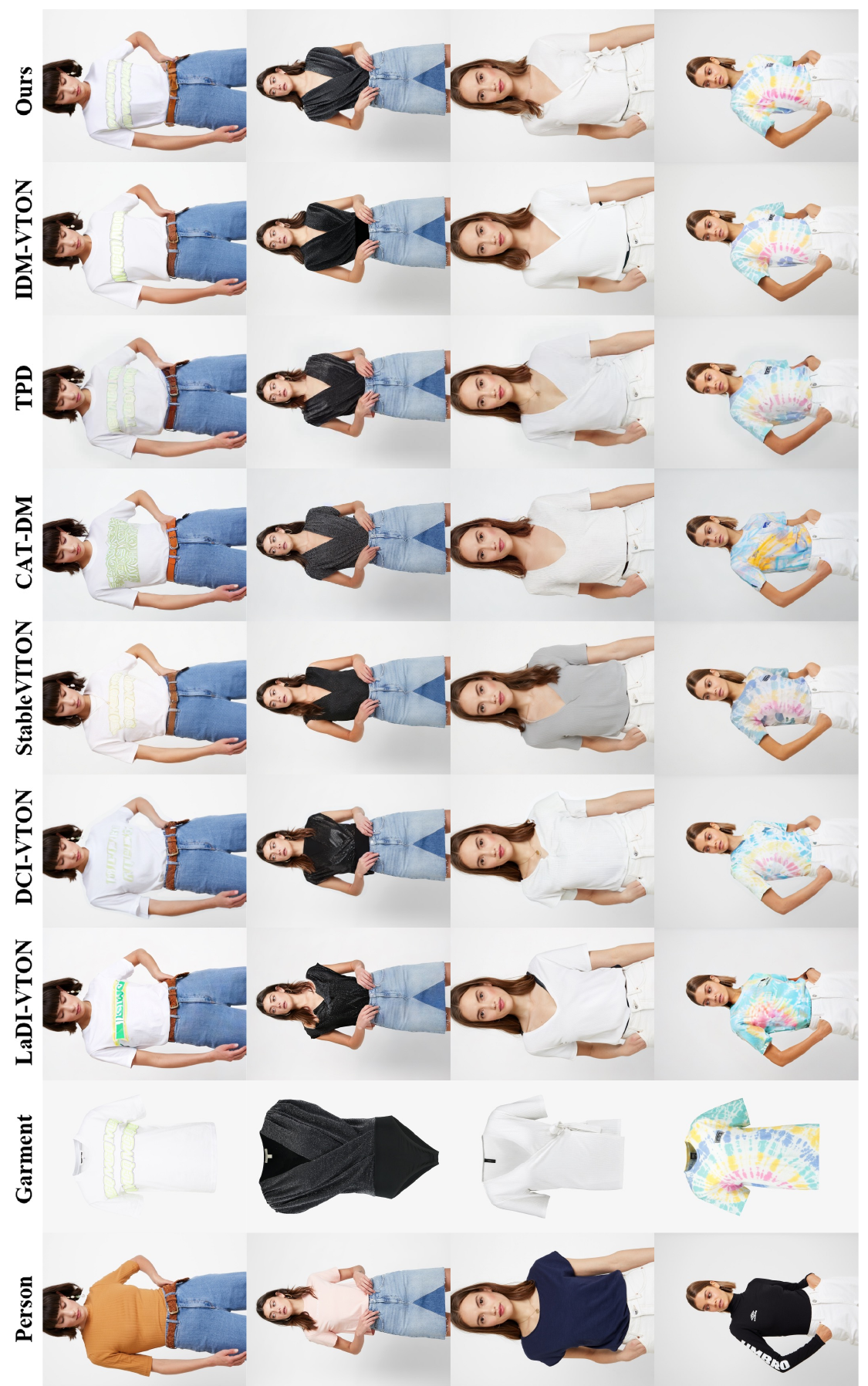}
   \caption{More qualitative comparison between our ITA-MDT and previous methods on the VITON-HD.}
\label{fig:qual3}
\end{figure*}

\begin{figure*}[t]
\centering
   \includegraphics[width=0.8\textwidth]{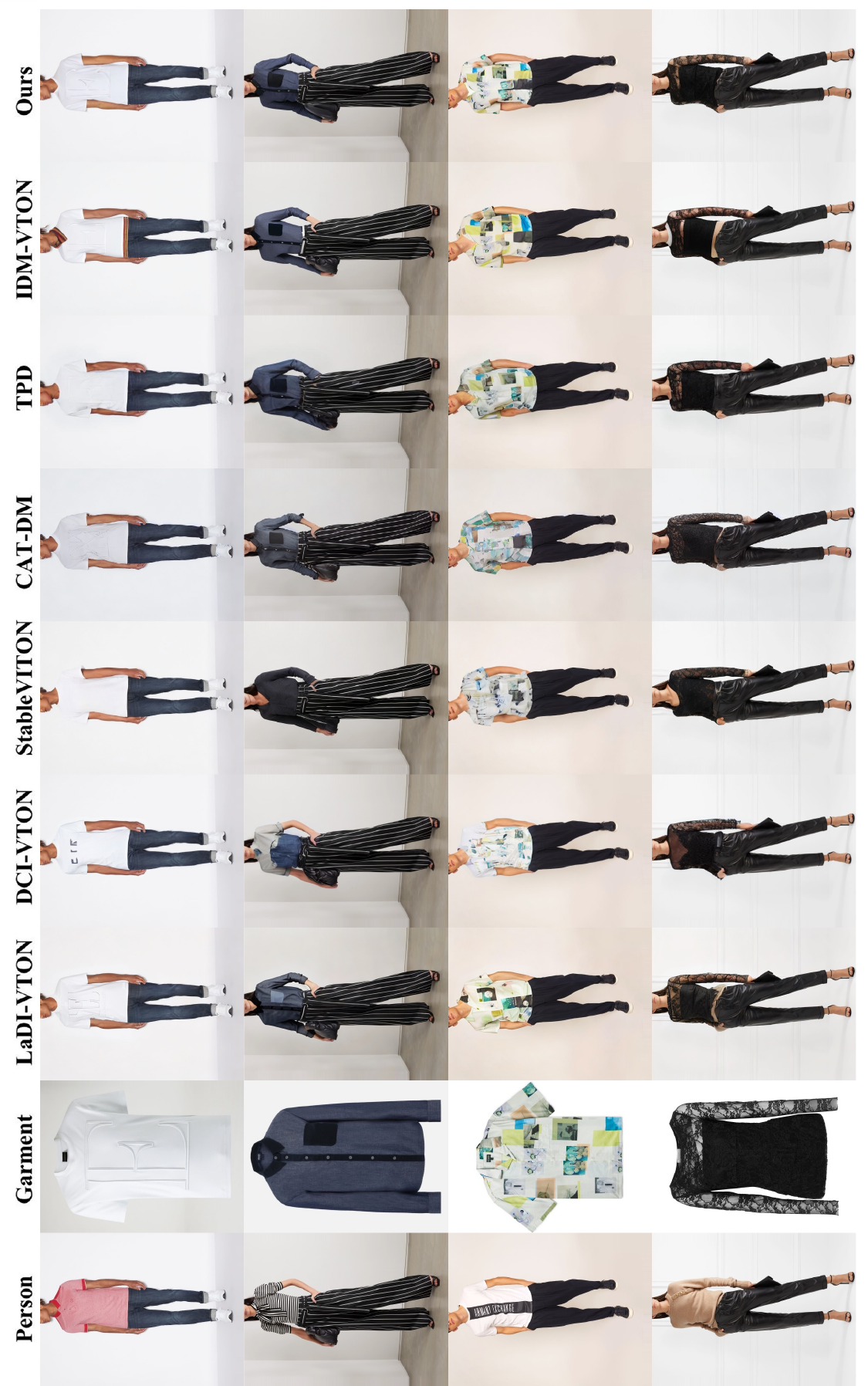}
   \caption{Qualitative comparison between our ITA-MDT and previous methods on the DressCode Upper-body.}
\label{fig:qual5}
\end{figure*}

\begin{figure*}[t]
\centering
   \includegraphics[width=0.8\textwidth]{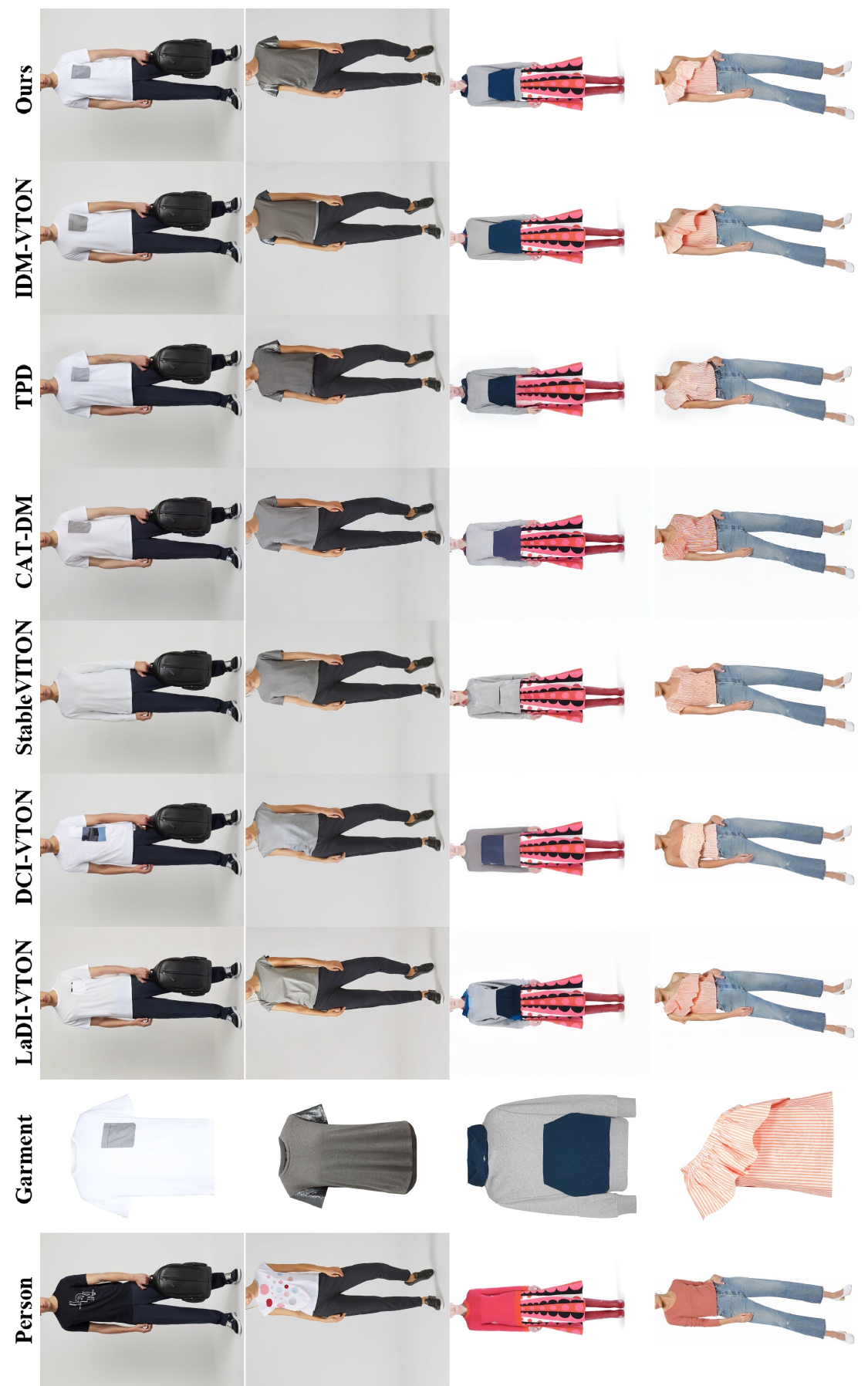}
   \caption{More qualitative comparison between our ITA-MDT and previous methods on the DressCode Upper-body.}
\label{fig:qual4}
\end{figure*}

\begin{figure*}[t]
\centering
   \includegraphics[width=1.0\textwidth]{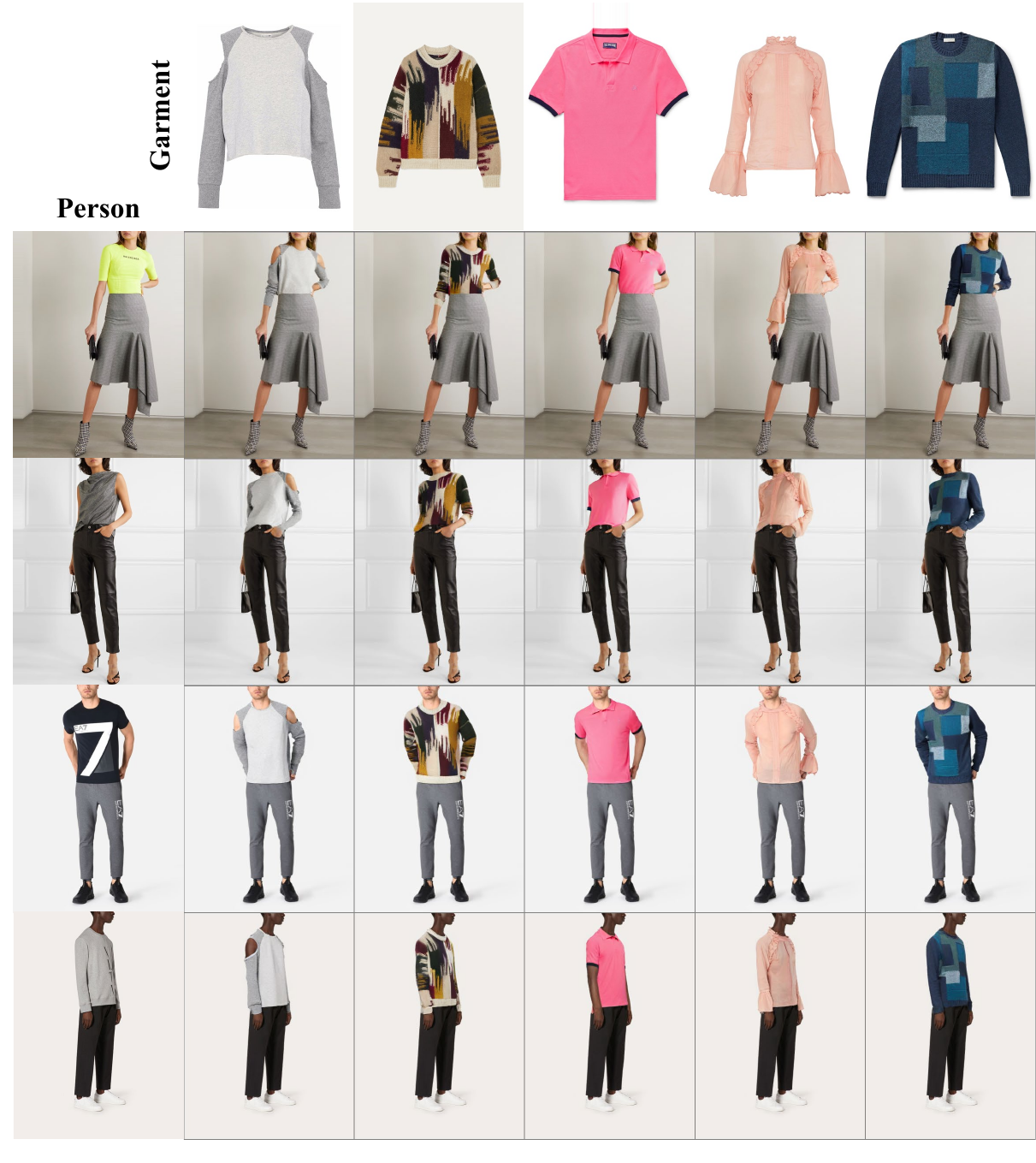}
   \caption{Qualitative results of our ITA-MDT on DressCode Upper-body.}
\label{fig:qual6}
\end{figure*}

\begin{figure*}[t]
\centering
   \includegraphics[width=1.0\textwidth]{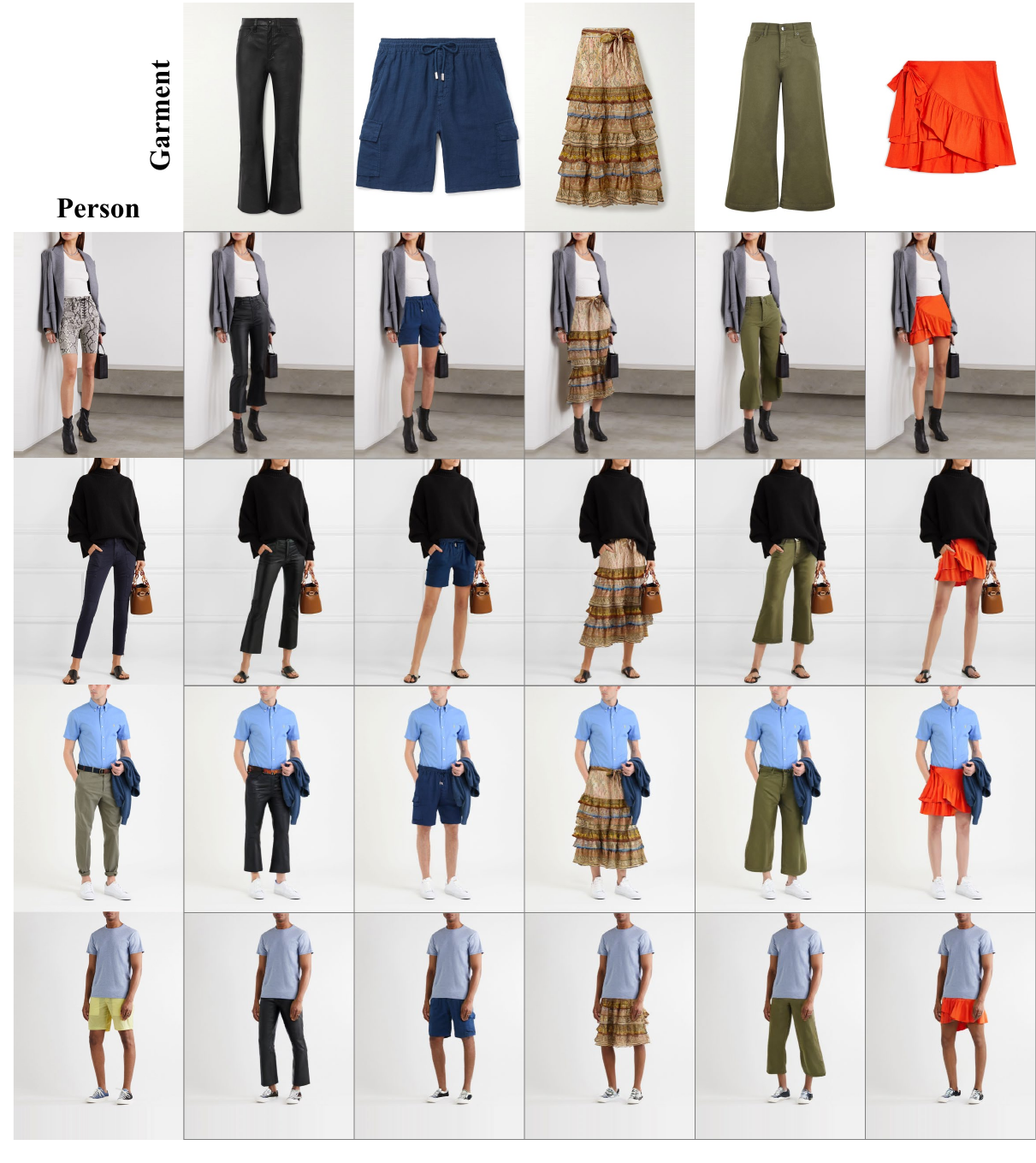}
   \caption{Qualitative results of our ITA-MDT on DressCode Lower-body.}
\label{fig:qual7}
\end{figure*}

\begin{figure*}[t]
\centering
   \includegraphics[width=1.0\textwidth]{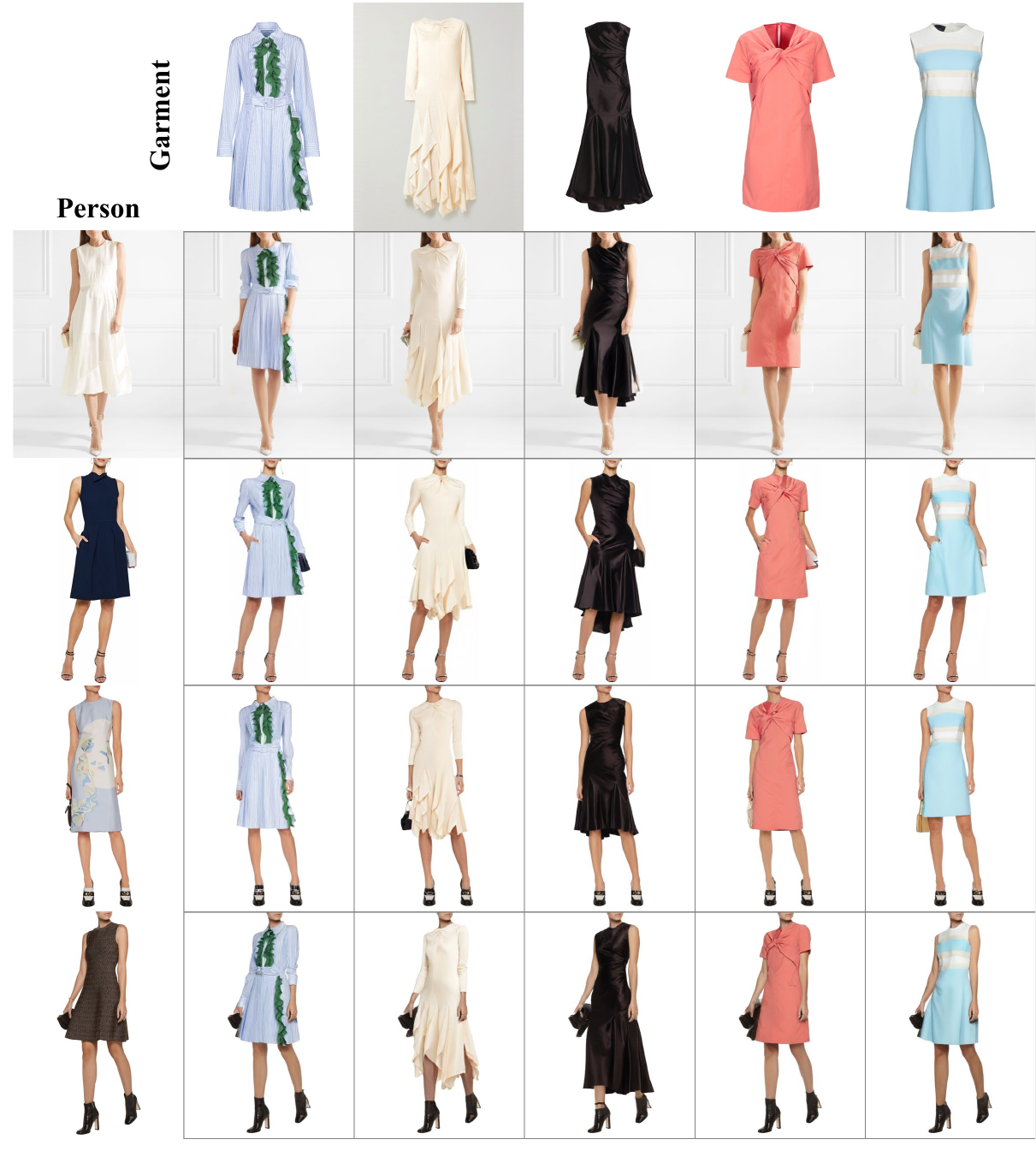}
   \caption{Qualitative results of our ITA-MDT on DressCode Dresses.}
\label{fig:qual8}
\end{figure*}

\begin{itemize}
    \item Figure \ref{fig:qual1}: Qualitative comparison of the effect of each component of the ITA-MDT framework on VITON-HD.
    \item Figure \ref{fig:qual2}: Qualitative comparison between our ITA-MDT and previous methods on the VITON-HD.
    \item Figure \ref{fig:qual3}: More qualitative comparison between our ITA-MDT and previous methods on the VITON-HD.
    \item Figure \ref{fig:qual5}: Qualitative comparison between our ITA-MDT and previous methods on the DressCode Upper-body.
    \item Figure \ref{fig:qual4}: More qualitative comparison between our ITA-MDT and previous methods on the DressCode Upper-body.
    \item Figure \ref{fig:qual6}: Qualitative results of our ITA-MDT on the DressCode Upper-body.
    \item Figure \ref{fig:qual7}: Qualitative results of our ITA-MDT on the DressCode Lower-body.
    \item Figure \ref{fig:qual8}: Qualitative results of our ITA-MDT on the DressCode Dresses.
\end{itemize}

\end{document}